\documentclass[sigconf]{acmart}
\AtBeginDocument{%
  }

\setcopyright{acmlicensed}
\copyrightyear{2026}
\acmYear{2026}
\acmDOI{XX.XXX}
\acmConference[Conference acronym 'XX]{Make sure to enter the correct
  conference title from your rights confirmation email}{June 03--05,
  2018}{Woodstock, NY}
\acmISBN{978-1-4503-XXXX-X/18/06}

\usepackage{pdfpages}
\usepackage{algorithmicx,algorithm}
\usepackage{algpseudocode}
\usepackage{algorithm,algpseudocode}
\usepackage{makecell,booktabs}
\usepackage{multirow}
\usepackage{multicol}  
\usepackage{bm}
\usepackage{siunitx}
\usepackage{xcolor}
\usepackage{enumitem}
\usepackage{subfigure}
\usepackage{xspace}
\usepackage{cleveref}
\usepackage{etoolbox}
\usepackage{pdfcomment}
\usepackage{amsmath}
\usepackage[normalem]{ulem}
\usepackage{marvosym}   
\usepackage{soul}
\usepackage{pifont}
\newcommand{\multirowoffset}{-0.5\dimexpr \aboverulesep + \belowrulesep + \cmidrulewidth}

\usepackage{tcolorbox}
\def\model{InsTraj\xspace}

\renewcommand\footnotetextcopyrightpermission[1]{}

\begin{document}

\title{InsTraj: Instructing Diffusion Models with Travel Intentions to Generate Real-world Trajectories}

\author{Yuanshao Zhu$^{1,2}$,~Yuxuan Liang$^{3}$,~Xiangyu Zhao$^{2}$, Liang Han$^{4}$,Xinwei Fang$^{5}$,\\~Xun Zhou$^{4}$,~Xuetao Wei$^{1}$,~James Jianqiao Yu$^{4}$}

\affiliation{
  \institution{$^1$Southern University of Science and Technology,~~$^2$City University of Hong Kong\\$^3$The Hong Kong University of Science and Technology (Guangzhou),\\$^4$Harbin Institute of Technology (Shenzhen),~~$^5$University of York}
  \country{}
  \address{}
\text{yuanshao@ieee.org}
}

\renewcommand{\shortauthors}{Yuanshao Zhu et al.}

\begin{abstract}
The generation of realistic and controllable GPS trajectories is a fundamental task for applications in urban planning, mobility simulation, and privacy-preserving data sharing.
However, existing methods face a two-fold challenge: they lack the deep semantic understanding to interpret complex user travel intent, and struggle to handle complex constraints while maintaining the realistic diversity inherent in human behavior.
To resolve this, we introduce \textbf{InsTraj}, a novel framework that instructs diffusion models to generate high-fidelity trajectories directly from natural language descriptions. 
Specifically, InsTraj first utilizes a powerful large language model to decipher unstructured travel intentions formed in natural language, thereby creating rich semantic blueprints and bridging the representation gap between intentions and trajectories.
Subsequently, we proposed a multimodal trajectory diffusion transformer that can integrate semantic guidance to generate high-fidelity and instruction-faithful trajectories that adhere to fine-grained user intent.
Comprehensive experiments on real-world datasets demonstrate that InsTraj significantly outperforms state-of-the-art methods in generating trajectories that are realistic, diverse, and semantically faithful to the input instructions.
\end{abstract}

\maketitle

\section{Introduction}
\begin{figure}[!t]
    \includegraphics[width=1\linewidth]{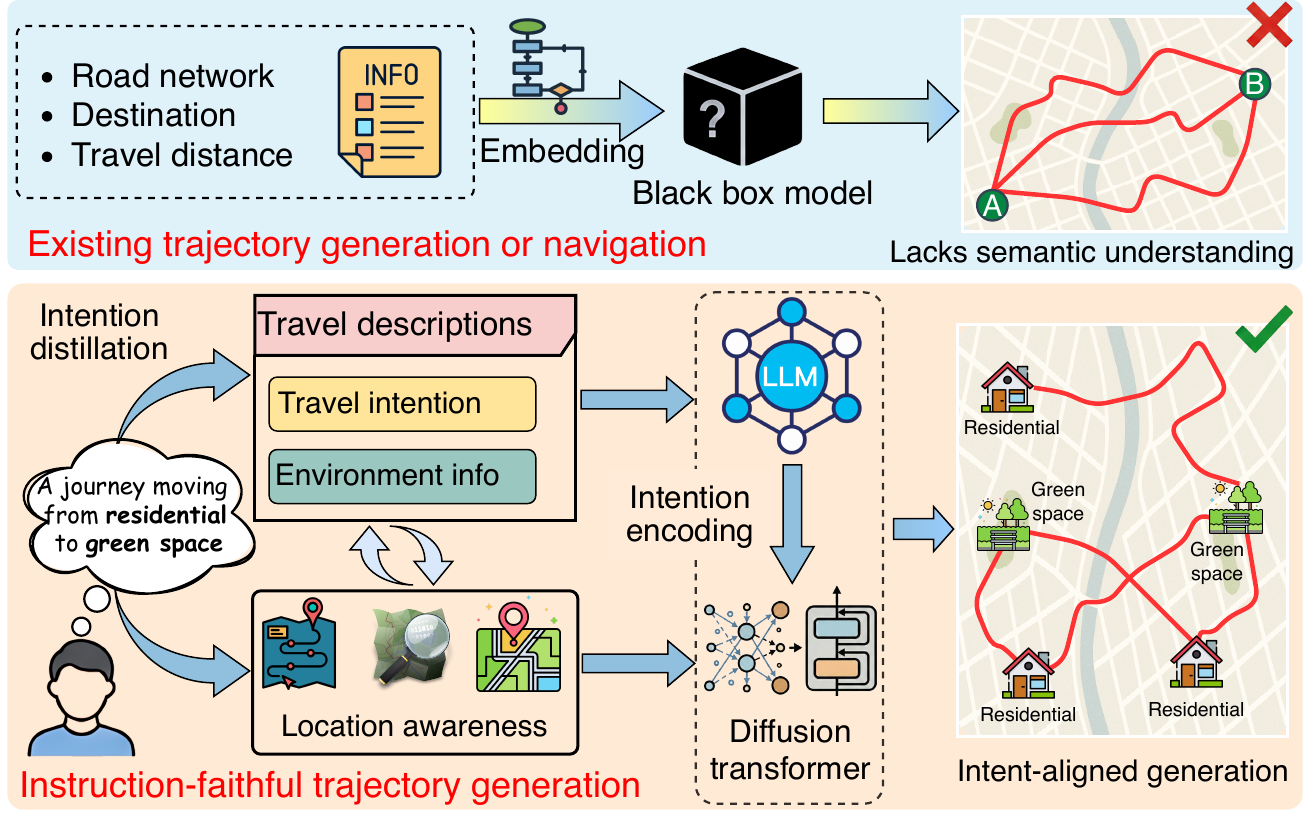}
    \Description[]{}
    \caption{Comparison of the proposed \model method with existing trajectory generation or navigation methods. The \model can effectively generate trajectories that are faithful to human travel intentions and exhibit diversity.
    }
    \label{fig:intro}
\end{figure}

The ubiquity of location-aware technologies, from smartphones to vehicle navigation systems, has resulted in an explosion of spatio-temporal trajectory data \cite{chen2024deep,guo2018learning}. 
This data provides insights into human mobility patterns and has become the cornerstone of numerous applications that shape modern society, from macro-level intelligent traffic management to micro-level personalized location services \cite{xu2024mm,guo2020context}.
Despite its value, large-scale trajectory corpora are difficult to acquire reliably: collection campaigns are expensive, coverage is often biased toward specific user groups, and raw trajectories cannot be shared freely because they carry sensitive personal information \cite{zhu2024controltraj}.
Meanwhile, navigation algorithms focus on finding deterministic optimal paths between coordinates and are unable to model the inherent diversity of human movement (shown in Figure \ref{fig:intro}).
Consequently, synthetic trajectory generation has emerged as a vital research area, providing realistic yet privacy-compliant substitutes for scarce real-world data \cite{peng2025diffusion}.

In pursuit of this target, deep generative models have shown tremendous potential in synthetic trajectory generation, such as variational autoencoders (VAEs) \cite{chen2021trajvae}, generative adversarial networks (GANs) \cite{cao2021generating}, and, more recently, diffusion models have emerged as leading solutions \cite{yang2024survey, peng2025diffusion}.
These methods excel at learning the underlying distributions, enabling them to reproduce point-wise geometry and aggregate flow statistics with striking accuracy. 
Still, the resulting outputs are inherently unguided and lack task-specific utility.
Furthermore, recent advancements have introduced controllability via physical constraints, such as start-and-end points or adherence to road networks \cite{zhu2024controltraj}. 
However, such control remains confined mainly to the geometric level (similar to traditional navigation tools that seek deterministic optimal paths), offering only coarse-grained manipulation while excluding the fine-grained semantic guidance required to model diverse human behaviors.

The above limitation exposes two core challenges that current methods are ill-equipped to handle:
1) \textbf{Semantic-spatio-temporal representation gap}: Existing models \cite{zhu2023difftraj} lack the ability to accurately comprehend the dependencies between human travel behavior and spatio-temporal dynamics, thereby failing to capture human abstract intentions (shown in Figure \ref{fig:intro}).
This results in a critical misalignment where trajectories, though geographically plausible, fail to embody the underlying travel purpose
2) \textbf{Generative control dilemma}: When subjected to complex constraints, models tend to either collapse into a single, deterministic path (thereby degenerating into a routing algorithm and sacrificing the inherent diversity of human behavior) or deviate from the target to preserve stochasticity \cite{zhu2024controltraj}. 
Consequently, bridging these gaps requires a framework that simultaneously interprets natural-language semantics and maintains flexible control over generation.

However, building such an instruction-faithful model is a non-trivial task. Specifically, it faces three challenges:
\begin{itemize}[leftmargin=*]
    \item \textbf{Intent distillation:} Human intentions are rarely explicitly logged, but are implicitly contained within initial GPS trajectories or free-form trip information. 
    Extracting and defining potential travel purposes (e.g., commuting, leisure) from semantically sparse trajectory data is the primary challenge.
    \item \textbf{Intent encoding:} Distilled intentions, often vague and unstructured, must be encoded into precise and continuous conditioning signals that deep generative models can accept.
    The challenge lies in creating representations that are semantically expressive yet granular enough to steer the generative process.
    \item \textbf{Intent-aligned generation:} Finally, a robust backbone network must be architected to generate high-fidelity trajectories from conditional signals.
    This requires ensuring strict adherence to every facet of the intent while simultaneously preserving the diversity of real-world behavior to avoid mode collapse.
\end{itemize}

To address these challenges, we propose \textbf{\model}, a novel framework that instructs diffusion models to generate realistic trajectories directly from travel intentions.
As illustrated in Figure \ref{fig:intro}, our approach systematically addresses each challenge by synergistically integrating Large Language Models (LLMs) and Multimodal Diffusion Transformers (DiT) \cite{peebles2023scalable}.
For intent distillation and encoding, we analyze the functional traits of the travel regions using Point-of-Interest (POI) data and regional semantics to automatically infer travel purposes and construct natural-language descriptions. 
\model then leverages a powerful LLM as its semantic engine, utilizing extensive world knowledge to interpret unstructured text and extract detailed travel intentions while encoding this understanding into high-dimensional semantic blueprints that serve as expressive conditioning signals for the subsequent generation stage.
For intent-aligned generation, the semantic blueprint steers our generative backbone: a multimodal trajectory DiT (\textbf{MT-DiT}). 
Its conditional modulation and joint attention mechanism ensure strict faithfulness by consistently aligning the iterative generation process with the blueprint, enforcing both global and local constraints. 
Together, the inherent stochasticity of diffusion enables the model to produce varied, plausible trajectories from the same instruction, preserving the essential nature of human behavior.

In summary, this work makes the following key contributions:
\begin{itemize}[leftmargin=*]
    \item We propose \model, a novel LLM-driven framework for generating instruction-faithful and semantically-guided GPS trajectories directly from human travel descriptions, addressing the critical gap between user intent and model output.
    \item We design a multimodal trajectory diffusion transformer that uniquely integrates LLM-derived semantic understanding into the generation process, enabling fine-grained control.
    \item We conduct comprehensive experimental validation on two real-world datasets, demonstrating the superiority of \model over state-of-the-art methods across metrics for realism, diversity, and, most critically, instruction faithfulness.
\end{itemize}

\section{Preliminary}\label{sec:pre}

\subsection{Definitions and Problem Formulation}
\noindent \textbf{Definition 1} \textbf{(Trajectory)}.
A GPS trajectory, denoted as $\boldsymbol{x}_0$, is defined as an ordered sequence of $n$ spatio-temporal points: $ \boldsymbol{x}_0= \{p_1, p_2, \dots, p_n\}$.
Each point $p_i$ is a tuple containing spatial coordinates, temporal information, and optional semantic attributes: $p_i = (\text{lat}_i, \text{lon}_i, t_i, \boldsymbol{s}_i)$.
Here, $\text{lat}_i$ and $\text{lon}_i$ denote the latitude and longitude coordinates, respectively, $t_i$ the timestamp satisfying the time ordering constraint.
$\boldsymbol{s}_i$ is an optional feature set used to profile the motion attributes of the entire trip, such as movement speed, duration time, and environmental conditions.

\noindent \textbf{Definition 2} (\textbf{Human Travel Intention}).
A Human travel instruction, denoted by $I$, is a natural language description provided by users that conveys their intent, preferences, or contextual constraints regarding a desired trajectory.
These instructions span from simple commands (e.g., ``go to the airport'') to complex, semantically rich narratives (e.g., ``I would like to take a walk in the park this afternoon, then go shopping and have dinner nearby'').

\noindent \textbf{Instruction-Faithful Trajectory Generation}. 
The primary objective of our work is to learn a conditional generative model, parameterized by $\theta$, that approximates the true data distribution $p(\boldsymbol{x}_0|I)$.
This model $p_{\theta}(\boldsymbol{x}_0|I)$ should be capable of synthesizing a realistic trajectory $\tilde{\boldsymbol{x}}_0 \sim p_{\theta}(\boldsymbol{x}_0|I)$  conditioned on a given human travel instruction $I$.  
Generally, the generated trajectory must satisfy:
\begin{itemize}[leftmargin=*]
    \item \textbf{Spatio-temporal Fidelity:} The trajectory $\tilde{\boldsymbol{x}}$ must be physically and statistically realistic. 
    It should exhibit movement patterns (e.g., speed, acceleration) and adhere to geographical constraints (e.g., road networks) consistent with real-world data.

    \item \textbf{Instruction Faithfulness:} The trajectory $\tilde{\boldsymbol{x}}$ must accurately embody all semantic cues and constraints specified in the instruction $I$, ensuring a strong alignment between the described intent and the generated output.

\end{itemize}

\subsection{Denoising Diffusion Probabilistic Model}
Denoising diffusion probabilistic models have emerged as a powerful class of generative models \cite{ddpm}, demonstrating remarkable success in synthesizing high-quality and diverse data samples \cite{diffusionsuvery}. 
Typically, it operates through two core processes: a fixed forward noising process and a learned reverse denoising process.

\noindent \textbf{Forward Noising Process}.
The forward process,  $q(\cdot)$, progressively injects Gaussian noise into an initial data sample $\boldsymbol{x}_0$ over $T$ discrete timesteps, according to a variance schedule $\{ \beta_t \in (0, 1)\}^{T}_{t=1}$.
A key property is that the noisy data at any step $t$, denoted $\boldsymbol{x}_t$, can be sampled in a closed form:
\begin{equation}
q(\boldsymbol{x}_t|\boldsymbol{x}_0) = \mathcal{N}(\boldsymbol{x}_t; \sqrt{\bar{\alpha}_t}\boldsymbol{x}_0, (1-\bar{\alpha}_t)\mathbf{I}),
\label{eq:forward_closed_form}
\end{equation}
where $\alpha_t = 1-\beta_t$ and $\bar{\alpha}_t = \prod_{s=1}^t \alpha_s$. As $t \to T$, $\boldsymbol{x}_T$ converges to a standard isotropic Gaussian distribution, effectively transforming the complex data distribution into a simple, tractable one.

\noindent \textbf{Reverse Denoising Process}.
The reverse process aims to learn this noising process in reverse, generating data by starting from pure noise $\boldsymbol{x}_T \sim \mathcal{N}(\boldsymbol{0}, \mathbf{I})$ and iteratively denoising it.
This process is also a Markov chain, modeled by parameterized with $\theta$, The goal is to approximate the true posterior $q(\boldsymbol{x}_{t-1} | \boldsymbol{x}_{t}, \boldsymbol{x}_{0})$ with a learned distribution  $p_{\theta}(\boldsymbol{x}_{t-1}|\boldsymbol{x}_t)$:
\begin{align}
   p_\theta(\boldsymbol{x}_{t-1}|\boldsymbol{x}_t) = \mathcal{N}(\boldsymbol{x}_{t-1}; \boldsymbol{\mu}_\theta(\boldsymbol{x}_t, t), \boldsymbol{\Sigma}_\theta(\boldsymbol{x}_t, t)).
\end{align}
For $\boldsymbol{\Sigma}_\theta$, it is typically set to a fixed value, such as $\beta_t \mathbf{I}$ or $\tilde{\beta}_t \mathbf{I} = \frac{1-\bar{\alpha}_{t-1}}{1-\bar{\alpha}_t}\beta_t \mathbf{I}$.
The mean $\boldsymbol{\mu}_\theta$ can then be reparameterized using this predicted noise:
\begin{align}
    \boldsymbol{\mu}_\theta(\boldsymbol{x}_t, t) = \frac{1}{\sqrt{\alpha_t}} \left( \boldsymbol{x}_t - \frac{\beta_t}{\sqrt{1-\bar{\alpha}_t}} \boldsymbol{\epsilon}_\theta(\boldsymbol{x}_t, t) \right).
\end{align}
This is achieved by training a neural network $\boldsymbol{\epsilon}_\theta$, typically parameterized as a U-Net or a Transformer, to predict the noise $\boldsymbol{\epsilon}$ that was added to $\boldsymbol{x}_0$ to obtain $\boldsymbol{x}_t$.

\noindent\textbf{Conditional Diffusion Models.}
To achieve instruction-faithful generation, we extend the unconditional DDPM to a \emph{conditional} model that learns the distribution $p_{\theta}(\boldsymbol{x}_0|\boldsymbol{c})$, where $\boldsymbol{c}$ is a conditioning vector derived from a human travel instruction. This is achieved by modifying the noise prediction network, $\boldsymbol{\epsilon}_\theta$, to accept $\boldsymbol{c}$ as an additional input. The network's new signature becomes $\boldsymbol{\epsilon}_\theta(\boldsymbol{x}_t, t, \boldsymbol{c})$, allowing the semantic information encoded in $\boldsymbol{c}$ to steer the denoising process at every step.
Consequently, the model is trained and optimized by minimizing a simplified mean-squared error objective with conditional signals:
\begin{equation}
\mathcal{L}= \mathbb{E}_{t, \boldsymbol{x}_0, \boldsymbol{c}} \left[ ||\boldsymbol{\epsilon} - \boldsymbol{\epsilon}_\theta(\sqrt{\bar{\alpha}_t}\boldsymbol{x}_0 + \sqrt{1-\bar{\alpha}_t}\boldsymbol{\epsilon}, t, \boldsymbol{c})||^2 \right].
\label{eq:loss_conditional}
\end{equation}

\noindent\textbf{Inference Procedure.} 
During inference, a new trajectory is generated by starting from pure noise $\tilde{\boldsymbol{x}}_T \sim \mathcal{N}(\boldsymbol{0}, \mathbf{I})$ and iteratively applying the learned denoising step, guided by the condition $\boldsymbol{c}$. For each step from $t=T, \dots, 1$, the iteratively model predicts the noise $\hat{\boldsymbol{\epsilon}} = \boldsymbol{\epsilon}_\theta(\tilde{\boldsymbol{x}}_t, t, \boldsymbol{c})$. 
This prediction is then used to compute the less-noisy sample $\tilde{\boldsymbol{x}}_{t-1}$ via the standard DDPM update rule, until the final clean sample $\tilde{\boldsymbol{x}}_0$ is obtained.

\section{Methodology}\label{sec:method}

\begin{figure*}[!t]
    \includegraphics[width=1\linewidth]{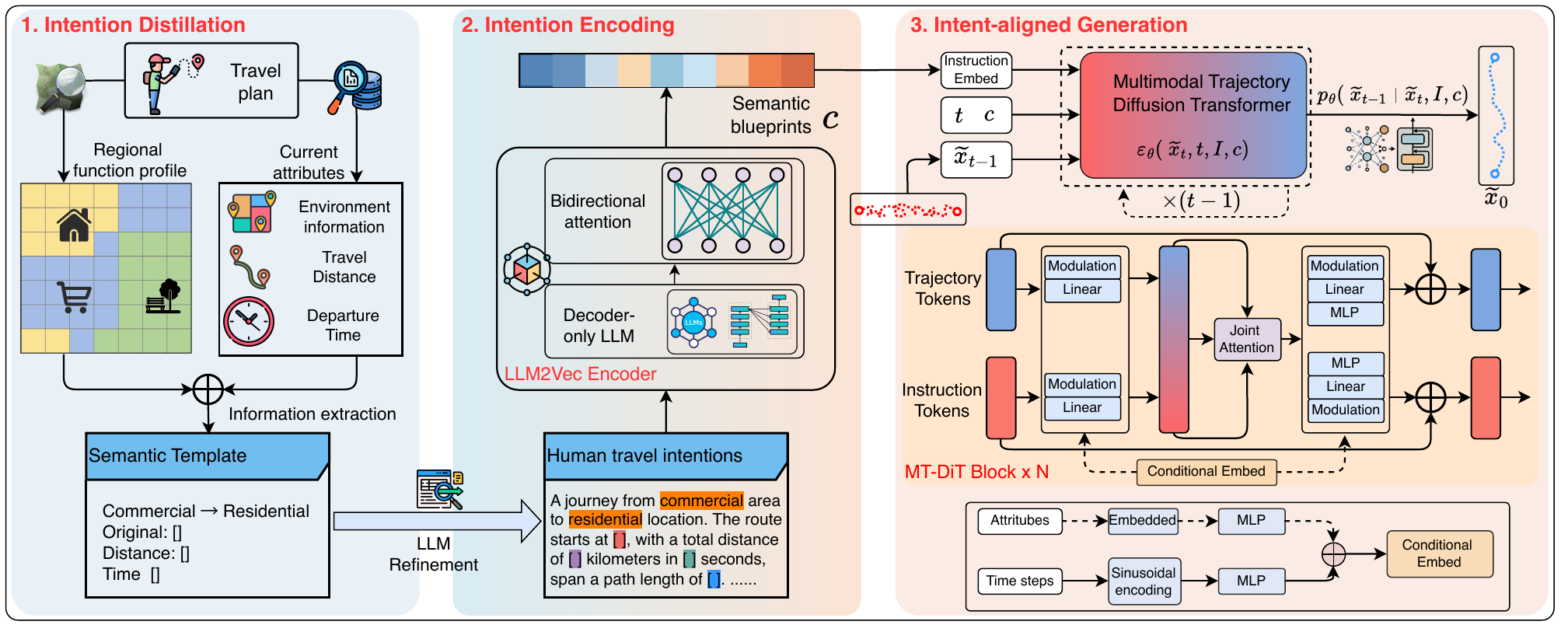}
    \Description[]{}
    \caption{
    Overview of the \model framework. It distills travel intentions from trajectory data, 
    encodes them via LLM into semantic blueprints, and employs 
    MT-DiT with Joint Attention to generate instruction-faithful trajectories.
    }
    \label{fig:overview}
\end{figure*}

\subsection{Overview of the Framework}
As illustrated in Figure \ref{fig:overview}, the proposed framework operates through three stages. 
First, we distill human travel intentions by leveraging functional regional analysis to infer latent mobility purposes from raw trajectory data.
Second, we encode these intentions by translating natural-language instructions into expressive, model-interpretable embeddings using a specially designed LLM.
Finally, we generate intent-aligned trajectories using our novel MT-DiT architecture, which fuses these semantic embeddings with spatio-temporal dynamics to ensure strict faithfulness to user intent.
Through the integration of these stages, \model directly resolves the fundamental challenges of constructing an instruction-faithful trajectory generator: \textbf{intent distillation, intent encoding, and intent-aligned generation}.

\subsection{Distilling Human Travel Intentions}\label{sec:distill_intention}

A primary obstacle to developing instruction-faithful models is the scarcity of datasets that pair real-world GPS trajectories with corresponding natural language descriptions. 
While existing methods often condition on simple, structured features like origin-destination (OD) pairs \cite{zhu2023difftraj, zhu2024controltraj}, such inputs are semantically impoverished and fail to capture the rich context behind a trip. 
To bridge this critical data gap, we propose a systematic pipeline to automatically extract and generate a large-scale corpus of instruction-trajectory pairs without manual annotation. 
Our pipeline is specifically designed to address two challenges: (1) \textbf{Semantic Grounding:} mapping abstract human intentions to observable spatio-temporal evidence;
and (2) \textbf{Instruction Prototyping:} converting the extracted semantics into diverse, natural language.

\subsubsection{\textbf{Regional Function Profiling.}}
To ground our instructions in real-world semantics, we first create a detailed functional map of the urban environment. 
The city space is partitioned into a fine-grained grid of uniform cells (e.g., 50m $\times$ 50m), balancing computational efficiency with semantic granularity.
Each grid cell $g_{i,j}$ is defined by its boundary latitude and longitude, constitutes the basic unit for analysis.
We leverage comprehensive Point-of-Interest (POI) data from OpenStreetMap to profile each cell. 
Specifically, all POIs within a cell are categorized into a predefined set of $K$ functional types (e.g., Residential, Commercial, Recreational) using a curated mapping schema based on OSM's hierarchical tagging system.
The functional profile of each cell $g_{i,j}$ is then quantified as a normalized vector $\boldsymbol{f}_{i,j} \in \mathbb{R}^{K}$. 
This is computed by first aggregating raw POI counts into a vector $\boldsymbol{v}_{i,j}$ and then applying  normalization:
$\boldsymbol{f}_{i,j} = \frac{\boldsymbol{v}_{i,j}}{\sum_{k=1}^{K} v_{i,j,k} + \epsilon}$,
where $\epsilon$ is a small smoothing constant. 
This probabilistic representation captures the mixed-use nature of urban areas while identifying the dominant functional characteristics of each region, effectively creating a rich, city-wide semantic map.

\subsubsection{\textbf{Trajectory Semantic Extraction.}}
With the city-wide functional map established, we proceed to infer the latent intent of each individual trajectory $\boldsymbol{x}_0$. 
This is a two-step process involving semantic feature extraction followed by natural language generation.
First, we extract a set of structured semantic features for each trajectory. 
We identify its origin ($p_1$) and destination ($p_n$) points.
Then, we retrieve their corresponding functional profile vectors, $\boldsymbol{f}_{org}$ and $\boldsymbol{f}_{dst}$, from our pre-computed map. 
The primary travel intent is inferred using a rule-based system that maps functional transitions to common activities (e.g., a Residential $\rightarrow$ Commercial transition is labeled as a ``shopping'' trip). 
In addition to this primary intent, we also extract key temporal attributes (e.g., time of travel) and travel style descriptors (e.g., travel distance, trajectory length) directly from the trajectory's spatio-temporal data.

Next, we convert this collection of structured semantic features into a natural language instruction. 
To ensure both consistency and diversity, we employ a template-based generation system augmented with LLM-based paraphrasing. 
We first populate a set of predefined templates with the extracted features to create a base instruction:
\textit{Functional transitions:} ``Travel from [origin type] area to [destination type] area''.
\textit{Purpose-driven:} ``Go [purpose] from [origin description] to [destination description]''.
\textit{Contextual:} ``Take a [time context] trip from [origin] to [destination]''.
To enhance linguistic naturalness and avoid repetitive phrasing, this template-generated sentence is then fed to a powerful LLM (e.g., GPT-4 \cite{achiam2023gpt}), which rephrases it into more varied and human-like expressions. 
Crucially, this LLM-refinement step introduces necessary linguistic diversity, preventing the downstream model from overfitting to rigid template patterns.

Please note that this pipeline is designed not only to generate instructions but also to implicitly define a spectrum of control: the more abstract and high-level an instruction is (e.g., ``go shopping''), the greater the potential for generative diversity.
Conversely, a highly specific instruction (e.g., specifying intermediate waypoints) would naturally constrain the generation to a narrower set of possibilities.
In summary, this systematic approach allows us to automatically generate a rich corpus of natural language descriptions that capture underlying mobility intentions, providing the foundation for training our instruction-faithful model.

\subsection{Encoding Semantic Intentions}\label{sec:LLM2vec}
Having established a corpus of instruction-trajectory pairs, we now address the challenge of intent encoding.
The core difficulty lies in translating unstructured natural language instructions into a continuous conditioning signal $\boldsymbol{c}$ capable of steering the diffusion process.
To bridge the semantic-spatio-temporal representation gap, we leverage the advanced semantic understanding capabilities of LLMs. 
While smaller encoders like BERT exist, the extensive world knowledge and instruction-following capabilities of modern LLMs are indispensable for interpreting complex spatial semantics.
However, standard LLMs (e.g., LLaMA \cite{touvron2023llama}) are optimized for generative dialogue rather than for representation learning, with training objectives misaligned with the semantic encoding required by our task.
Their decoder-only architecture uses causal attention, which restricts a token's view to only its preceding context, preventing the formation of globally contextualized representations essential for understanding holistic travel instructions.
Although smaller bidirectional encoders like BERT \cite{devlin2019bert} can provide global contextualization, they lack the extensive world knowledge and instruction-following capabilities that are critical for interpreting complex spatial semantics. 
In addition, developing a dedicated bidirectional encoder from scratch would be computationally prohibitive and would forgo the rich semantic priors already embedded in modern LLMs.

\subsubsection{\textbf{LLM2Vec.}}
To resolve this, we adopt an unsupervised method that transforms any pre-trained decoder-only LLM into a powerful, bidirectional text encoder, namely LM2Vec \cite{llm2vec}. 
Specifically, this method first \textit{ enables bidirectional attention} by modifying the model's attention mask with an all-ones matrix, thus enabling full contextual understanding. 
It then adapts the LLM to this new capability through a \textit{Masked Next Token Prediction} fine-tuning objective to predict masked words within a description based on both past and future context.
Finally, the quality of the sequence-level embeddings is further refined using an \textit{unsupervised contrastive learning} stage based on SimCSE \cite{gao2021simcse} to maximize similarity between different augmented instructions while minimizing similarity with other instructions in the batch.
Upon completion of this three-stage process, our fine-tuned LLM, $E_\phi$, functions as a powerful instruction encoder. 
It transforms a given instruction $I$ into a sequence of token-level embeddings, from which we derive the final conditioning signal $\boldsymbol{c}$. 
This signal robustly captures both explicit constraints and implicit preferences, providing precise guidance for the final synthesis stage.

\subsubsection{\textbf{TrajCLIP}}
Quantitatively evaluating instruction faithfulness poses a significant challenge: there is no universal metric for scoring the semantic alignment between a generated trajectory and a free-form instruction. 
To address this, we propose TrajCLIP, a dedicated multi-modal model trained to measure this alignment.
Inspired by the success of CLIP \cite{radford2021learning} in aligning images and text, TrajCLIP consists of two parallel encoders: a Transformer-based trajectory encoder $\boldsymbol{v}_{\text{traj}}$ that processes sequences of trajectory points, and our fine-tuned LLM2Vec text encoder $\boldsymbol{v}_{\text{text}}$. 
We train these two encoders from scratch on our large-scale corpus of instruction-trajectory pairs using a symmetric contrastive loss. 
During training, the model learns to maximize the cosine similarity of genuine (trajectory, text) pairs and minimize the mismatched pairs:
\begin{equation}
\mathcal{L}_{\text{traj}\to\text{text}} = -\sum_{i=1}^{N} \log \frac{\exp(\text{sim}(\boldsymbol{v}_{\text{traj}}^i, \boldsymbol{v}_{\text{text}}^{i}) / \tau)}{\sum_{j=1}^{N} \exp(\text{sim}(\boldsymbol{v}_{\text{traj}}^j, \boldsymbol{v}_{\text{text}}^{j}) / \tau)},
\end{equation}
where $\text{sim}(\cdot,\cdot)$ is the cosine similarity and $\tau$ is a learnable temperature parameter. 
The final training objective is a symmetric loss computed over both directions. 
Once trained, the frozen TrajCLIP model serves as an effective, unbiased evaluator. 
It computes the cosine similarity between any given trajectory and instruction, a score we term the Trajectory-Instruction Similarity (TIS), which serves as a quantitative measure for our instruction faithfulness experiments \textbf{(RQ2) in Section \ref{sec:exp_rq2}}.

\subsection{Generating Intent-Aligned Trajectories}
With a powerful mechanism for encoding instructions into semantic embeddings, the final and most critical stage is to synthesize a trajectory that faithfully adheres to this guidance. 
The fundamental challenge lies in effectively fusing the high-level, abstract information from the instruction embedding with the low-level, iterative denoising process of the diffusion model. 
This fusion must be deep and continuous, steering the generation at every step, and computationally efficient, as trajectory generation can involve long sequences and numerous steps. 
To address both these needs, we depart from traditional U-Net backbones and propose the Multi-modal Trajectory Diffusion Transformer (MT-DiT). 
This novel architecture leverages the scalability of Transformers while being specifically designed for the deep, multi-layer fusion of textual and spatio-temporal data streams via joint attention.

\subsubsection{\textbf{MT-DiT Architecture}}
Our generative backbone MT-DiT is a pure Transformer architecture. 
It processes two different token sequences, one for trajectories and one for instructions, and is modulated by external information through a dedicated block design.
Inspired by recent advancements in multi-modal generative architectures \cite{esser2024scaling}, we process both modalities as token sequences that interact deeply within the network.

\noindent \textbf{Input Tokenization.} 
Firstly, all inputs are converted into token sequences in a shared hidden dimension $d$:
\begin{itemize}[leftmargin=*]
    \item \textbf{Trajectory Tokens: } A noisy trajectory $\boldsymbol{x}_t \in \mathbb{R}^{L \times 2}$ (where $L$ is the sequence length), is partitioned into non-overlapping ``patches'' of size $P$. 
    These patches are flattened and linearly projected into trajectory tokens, which we then add sinusoidal positional encodings to.
    This strategy allows the model to operate on coherent local patterns while significantly reducing the computational cost (from $L$ to $\left \lfloor   L/P\right \rfloor$) of subsequent modules.
    
    \item \textbf{Instruction Tokens:} The textual instruction $I$ is encoded by our LLM2Vec encoder into a sequence of token-level embeddings, which are then projected to dimension $d$ using linear layers.
    
    \item \textbf{Timestep and Attributes Embedding}: The diffusion time step $t$ and optional attributes are converted to vector embeddings using the standard practice described in \cite{zhu2023difftraj}.
\end{itemize}

\noindent \textbf{Joint Attention Mechanism.}
The core of our MT-DiT is a stack of $N$ identical blocks, each designed to facilitate deep inter-modal fusion. 
Within each block, the trajectory ($\boldsymbol{z}_{\text{traj}}$) and instruction ($\boldsymbol{z}_{\text{inst}}$) tokens are processed through distinct pathways but interact via a shared Joint Attention mechanism, inspired by recent multi-modal architectures \cite{esser2024scaling}.
Unlike standard cross-attention that only allows unidirectional conditioning, Joint Attention concatenates sequences to enable bidirectional flow.
Specifically, the mechanism first projects each modality into Query, Key, and Value representations using modality-specific weight matrices. 
This allows the model to learn distinct projection patterns for trajectory ($\mathbf{Q}_{\text{traj}}, \mathbf{K}_{\text{traj}}, \mathbf{V}_{\text{traj}}$) and text ($\mathbf{Q}_{\text{instr}}, \mathbf{K}_{\text{instr}}, \mathbf{V}_{\text{instr}}$) data.
The resulting sequences from both modalities are then concatenated along the sequence dimension (e.g., $\mathbf{Q}=[\mathbf{Q}_{\text{traj}},\mathbf{Q}_{\text{instr}}]$), 
and fed into a single multi-head scaled dot-product attention function. 
This step enables deep information exchange between the two modalities, allowing trajectory tokens to directly attend to specific semantic instruction tokens. 
Following this fusion, the output sequence is split back into its respective streams, each processed by a separate Feed-Forward Network. This design elegantly combines shared fusion with modality-specific processing.

\noindent\textbf{Conditional Modulation via adaLN-Zero.}
To inject the conditional information (primarily the diffusion timestep $t$), we employ the adaLN-Zero mechanism \cite{peebles2023scalable}. 
This technique enhances standard Adaptive Layer Normalization (adaLN) by incorporating a learned scaling factor $\alpha$ directly into the residual path of each sub-layer (e.g., Joint Attention, FFN).
For a given hidden state $\boldsymbol{h}$, the output of a sub-layer using adaLN-Zero is computed as:
\begin{align}
    \text{output} &= \boldsymbol{h} + \alpha \cdot \text{SubLayer}(\text{adaLN}(\boldsymbol{h})) \\
    \text{adaLN}(\boldsymbol{h}) &= {\gamma} \odot \text{LayerNorm}(\boldsymbol{h}) + {\beta}
\end{align}
A small MLP dynamically generates all three conditioning parameters (i.e., the scale $\gamma$, the shift $\beta$, and the residual scaling factor $\alpha$) from the timestep embedding.
Crucially, initializing $\alpha=0$  makes each Transformer block initially behave as an identity function, a technique known to significantly stabilize training and improve final model performance.
Additionally, the derived $\gamma$ and $\beta$ enable time-based concepts to systematically influence the entire denoising process, thereby learning corresponding congestion patterns and activity area distributions.

\noindent\textbf{Final Layer.}
After the final MT-DiT block, the instruction tokens are discarded. The processed trajectory tokens are passed through a final linear layer that projects them back to the trajectory patch dimension, yielding the noise prediction  $\boldsymbol{\epsilon}_\theta(\tilde{\boldsymbol{x}}_t, t, \boldsymbol{c})$.

\section{Experiments}\label{sec:exper}
In this section, we first detail our experimental setup, encompassing datasets, implementation specifics, baselines, and evaluation metrics. 
Then, we evaluate \model via extensive experiments to address four key research questions:
\begin{itemize}[leftmargin=*]
    \item \textbf{RQ1 (Spatio-Temporal Fidelity)}: Does \model match or exceed existing baselines in capturing real-world mobility distributions?
    \item \textbf{RQ2 (Instruction Faithfulness)}: Can the proposed model generate trajectories consistent with human travel intentions?
    \item \textbf{RQ3 (Efficiency and Performance Analysis)}: What is the computational performance and efficiency profile of \model?
    \item \textbf{RQ4 (Spatio-Temporal Discovery)}: Beyond generation, can \model serve as a tool for spatio-temporal analysis and discovering mobility patterns through case studies?
\end{itemize}

\subsection{Experimental Setups}

\noindent {\textbf{Datasets.}}
We utilize two large-scale, real-world datasets commonly used in trajectory generation: \textit{Taxi-Chengdu} and \textit{Taxi-Xi'an} \cite{chen2024deep}.
Following standard pre-processing \cite{zhu2023difftraj}, we filter sparse trajectories and remove duplicates, reserving 50,000 samples for testing. 
Regional functional attributes (discussed in Section \ref{sec:distill_intention}) are derived from OpenStreetMap (OSM) POI data. 

\noindent {\textbf{Baselines.}}
We compare \model against representative baselines spanning different generative modeling paradigms, from classic GANs to the most recent controllable diffusion models.
Such as VAE \cite{chen2021trajvae}, TrajGAN \cite{liu2018trajgans}, DP-TrajGAN \cite{zhang2022dp, rao2020lstm}, Diffwave \cite{kongdiffwave}, DiffTraj \cite{zhu2023difftraj}, and ControlTraj \cite{zhu2024controltraj}.
Crucially, to ensure a fair comparison for instruction faithfulness (RQ2), we augmented the conditional modules of DiffTraj and ControlTraj with our LLM-derived intention embeddings. 
For spatio-temporal fidelity (RQ1) and efficiency (RQ3), we retained their original, unmodified versions to evaluate native performance. Baseline details are in \textbf{Appendix \ref{app:baseline}}.

\noindent {\textbf{Evaluation Metrics.}}
We employ two metric categories (their definitions are detailed in \textbf{Appendix \ref{app:metrics}}):
\begin{itemize}[leftmargin=*]
    \item \textbf{Spatio-Temporal Fidelity:} Following \cite{zhu2023difftraj,zhu2024controltraj}, we use Jensen-Shannon Divergence (JSD) to measure the distributional similarity between generated and real trajectories across \textbf{density, trip distribution, and travel length}.
    
    \item \textbf{Instruction Faithfulness:} We propose the Functional Consistency Rate (\textbf{FCR}) to verify if start/destination points match the instructed functional areas. 
    Additionally, we use the Trajectory-Instruction Similarity (\textbf{TIS}) score, computed by our TrajCLIP model (Sec. \ref{sec:LLM2vec}), to quantify global semantic alignment.
\end{itemize}
Further details of all metrics are provided in \textbf{Appendix \ref{app:metrics}}.

\noindent {\textbf{Implementation Details.}}
\model is implemented in PyTorch and involves specific architectural choices for the MT-DiT backbone and various hyperparameter settings. 
For a detailed breakdown of the model configuration, please see \textbf{Appendix \ref{app:prars_setting}}.

\subsection{Spatio-Temporal Fidelity (RQ1)}

\begin{table}[t]
\centering
\small
\caption{Quantitative comparison of spatio-temporal fidelity.}
\begin{tabular}{l|lccc}
\toprule
\textbf{Dataset} & \textbf{Method} & \textbf{ Density $\downarrow$} & \textbf{Trip $\downarrow$} & \textbf{Length $\downarrow$}   \\ 
\cmidrule(lr){1-5} 
\multirow{7}{*}{Chengdu} 
& VAE & 0.0139 & 0.0502   & 0.0368    \\
& TrajGAN  & 0.0137	& 0.0488 & 0.0329  \\
& DP-TrajGAN  & 0.0127 &	0.0438	& 0.0234  \\
& Diffwave  & 0.0145 & 0.0253 & 0.0315 \\
& DiffTraj   & 0.0055 & 0.0154 & 0.0169  \\
& ControlTraj  & 0.0039 & 0.0106 & 0.0117  \\
& \model  & \textbf{0.0032} & \textbf{0.0104} & \textbf{0.0094}\\
\cmidrule(lr){1-5} 
\multirow{7}{*}{Xi'an} 
& VAE   & 0.0237 & 0.0608 & 0.0497  \\
& TrajGAN  & 0.0220 & 0.0512  & 0.0386  \\
& DP-TrajGAN  & 0.0207 & 0.0498 & 0.0436   \\
& Diffwave   &  0.0213 & 0.0343 & 0.0321  \\
& DiffTraj    & 0.0126  & 0.0165 & 0.0203  \\
& ControlTraj  &  0.0104 & 0.0125 & 0.0168  \\
& \model  & \textbf{0.0093} & \textbf{0.0110} & \textbf{0.0131}   \\
\bottomrule
\end{tabular}
\label{tab:comp}
\end{table}

Table \ref{tab:comp} reports the quantitative performance of \model against baseline methods.
\model consistently achieves state-of-the-art performance across all metrics on both datasets, significantly outperforming the strongest competitor, ControlTraj.
Specifically, on the Chengdu dataset, \model reduces the JSD for Spatial Density from 0.0039 to 0.0032 and achieves significant gains in alignment of the Trip and Length distributions.
These improvements validate our architectural choices:
(1) The superior Density scores suggest that incorporating high-level semantic intentions allows the model to better capture the underlying purpose of mobility, leading to more realistic spatial distributions than purely geometric constraints.
(2) The improvements in Trip and Length metrics demonstrate that our MT-DiT backbone effectively models long-range spatio-temporal dependencies, preventing the structural fragmentation often observed in standard U-Net-based diffusion models.
This significant difference suggests that the model learns not just \textit{where} movement happens, but also \textit{how} and \textit{why}, leading to a more faithful reflection of the true data distribution.  

\begin{figure}[t]
    \subfigure[Generated Trajectory.]{
    \includegraphics[width=0.44\linewidth]{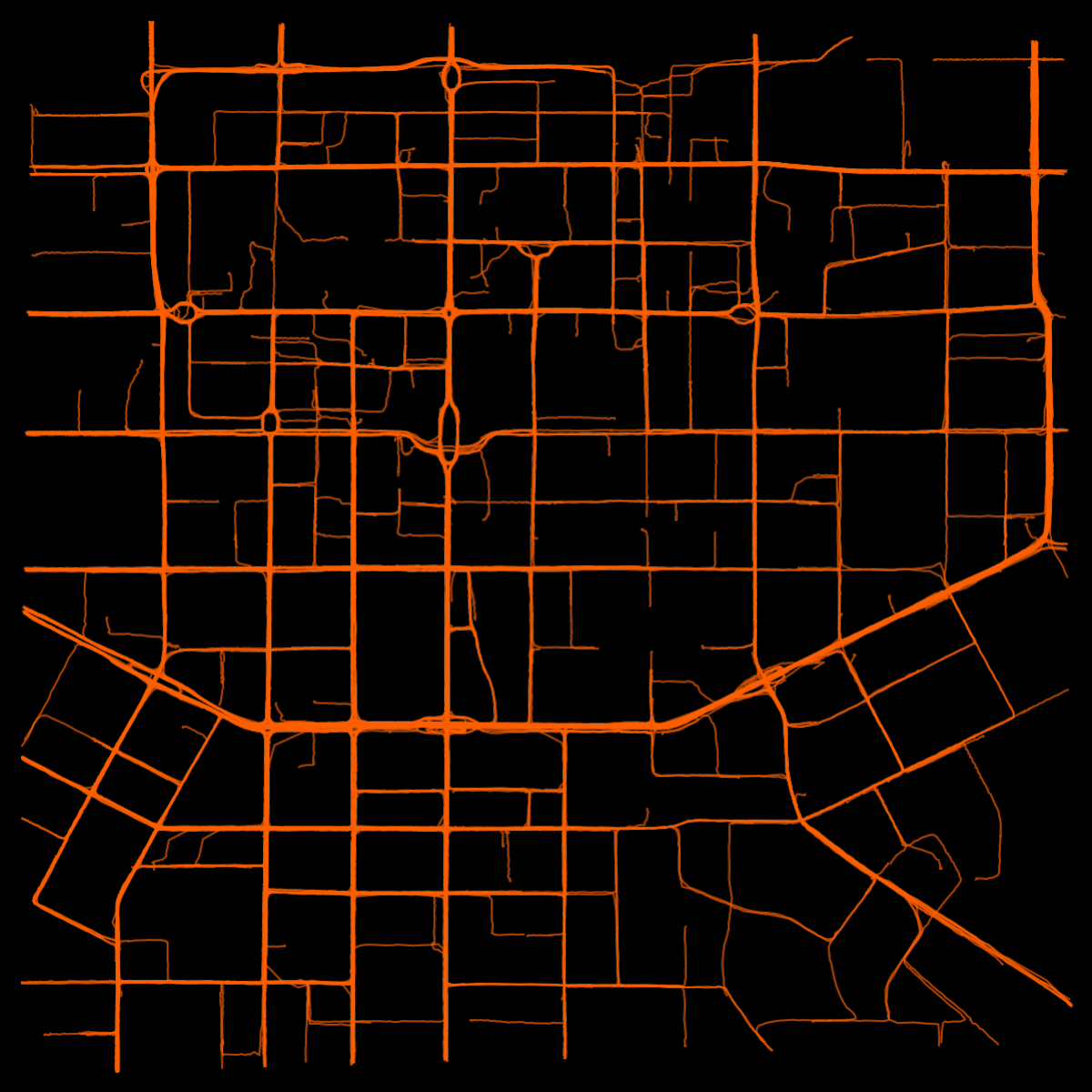}
    } \hspace{-0.025\linewidth} 
    \subfigure[Travel distance distribution (km).]{
    \includegraphics[width=0.52\linewidth]{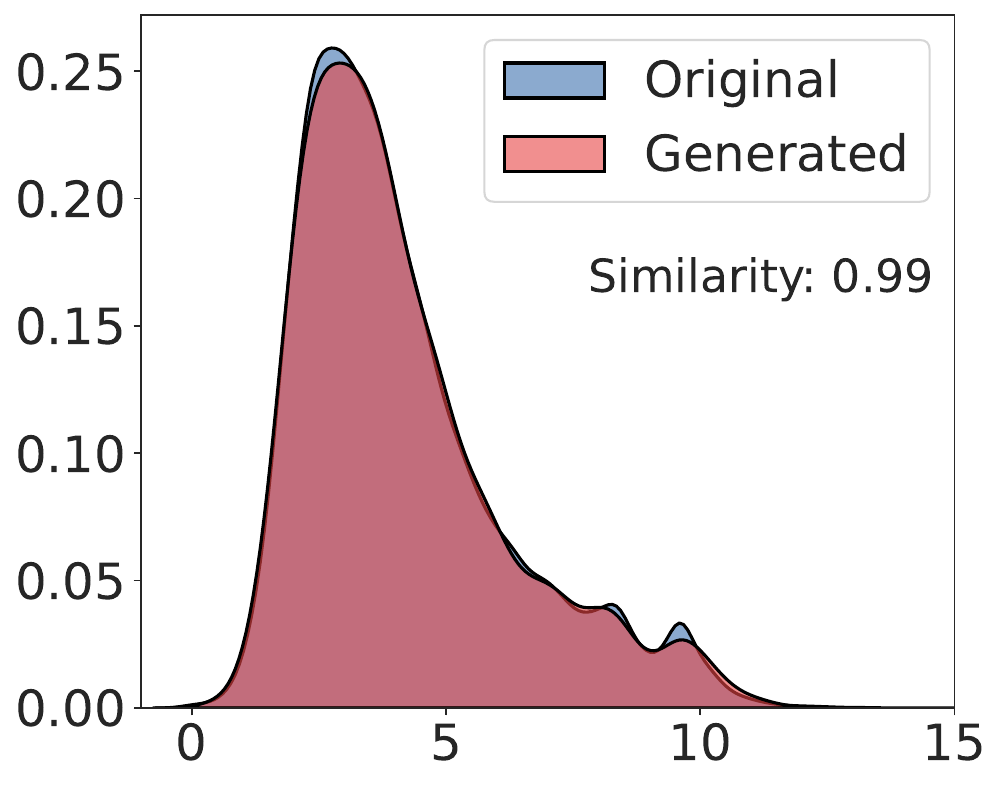}
     }
    \caption{Visualization of the geographical distribution and attributes of generated trajectories.}
        \Description[]{}
    \label{fig:traj_vis_xa}
\end{figure}

To provide intuitive validation of these quantitative findings, Figure \ref{fig:traj_vis_xa} presents a visualization of the generation results in Xi'an. 
From a macro-level perspective (Figure \ref{fig:traj_vis_xa}(a)), the trajectories generated by \model accurately reconstruct the city's skeletal structure, exhibiting high density on arterial roads and underlying geography in local streets.
From a micro-level statistical perspective (Figure \ref{fig:traj_vis_xa}(b)), the distribution of travel distances generated by \model overlaps almost perfectly with the ground truth (Similarity Score $>$ 0.99).
Taken together, these visualizations affirm that \model excels at capturing both the complex geographical constraints of an urban environment and the fundamental statistical properties of real-world mobility.

\subsection{Instruction Faithfulness (RQ2)}\label{sec:exp_rq2}
We now address the central question of our work: \textit{Can the model generate trajectories that strictly adhere to human travel intentions?} 
To answer this, we evaluate the alignment between instructions and generated trajectories using FCR-S or FCR-D (checking specific Start/Destination constraints) and TIS score (measuring global semantic consistency). 
In addition, we further analyzed the effects of instruction quality and granularity, with the results presented in \textbf{Appendix \ref{app:ins_ana}}.

\begin{table}[h]
\centering
\small
\caption{Analysis of instruction faithfulness and the contribution of key components.}
\begin{tabular}{l|lccc}
\toprule
\multirow{2}{*}[\multirowoffset]{\textbf{Dataset}} & \multirow{2}{*}[\multirowoffset]{\textbf{Method}} & \multicolumn{2}{c}{\textbf{FCR(\%) $\uparrow$}} & \multirow{2}{*}[\multirowoffset]{\textbf{TIS $\uparrow$}} \\
\cmidrule(lr){3-4}
& & \textbf{S}  & \textbf{D} &   \\
\cmidrule(lr){1-5} 
\multirow{5}{*}{Chengdu} 
& DiffTraj &  75.14 & 74.81  & 0.351\\
& ControlTraj & 75.54 & 74.61  & 0.350 \\
&  \model(Bert) & 75.40 & 75.62  & 0.348  \\
&  \model(DiT) & 82.83 & 81.37 & 0.369  \\
& \model &  \textbf{83.03}  & \textbf{82.54}  & \textbf{0.371 }   \\

\cmidrule(lr){1-5} 
\multirow{5}{*}{Xi'an} 
& DiffTraj &  84.75 &  84.69  &   0.385 \\
& ControlTraj & 85.49 & 84.73  & 0.386 \\
&  \model(Bert) & 85.76 & 85.68  & 0.385  \\
&  \model(DiT) & 89.51 & 89.29  & 0.411  \\
& \model &  \textbf{90.34}  & \textbf{89.93}  & \textbf{0.415}  \\
\bottomrule
\end{tabular}
\label{tab:intenion_comp}
\end{table}

\noindent\textbf{Superiority of \model.}
The main results are presented in Table \ref{tab:intenion_comp}.
We can easily observe that \model achieves a dominant performance, surpassing the strongest baseline (augmented ControlTraj) by a significant margin (e.g., $+7.5\%$ FCR-S and $+7.9\%$ FCR-D on Chengdu).
Recall that for fairness, the baselines were augmented with our LLM-derived embeddings. 
Thus, their lower performance highlights a critical insight: merely conditioning on semantic embeddings is insufficient. 
The superior performance of \model stems from the MT-DiT architecture, which enables deep, token-level fusion of these embeddings into the generative process, preventing the semantic signal from being diluted during diffusion.

\noindent\textbf{Impact of Model Components.} Table \ref{tab:intenion_comp} also dissects the contribution of our core components: (1) \textbf{Encoder Importance:} \model(Bert) exhibits a drastic performance drop, degrading to the level of standard baselines. 
This confirms that a generic language encoder (BERT) lacks the spatial semantic reasoning capabilities of our fine-tuned LLM2Vec, which is indispensable for interpreting complex travel instructions. 
(2) \textbf{Fusion Importance:} \model(DiT), which uses the correct encoder but a simpler DiT backbone, still underperforms the full model. 
This isolates the contribution of our Joint Attention mechanism, proving that bidirectional multi-modal fusion is architecturally superior for integrating textual guidance compared to simple concatenation or cross-attention.

\begin{figure}[t]
\centering 
    \subfigure[Efficiency comparison.]{
        \includegraphics[width=0.465\linewidth]{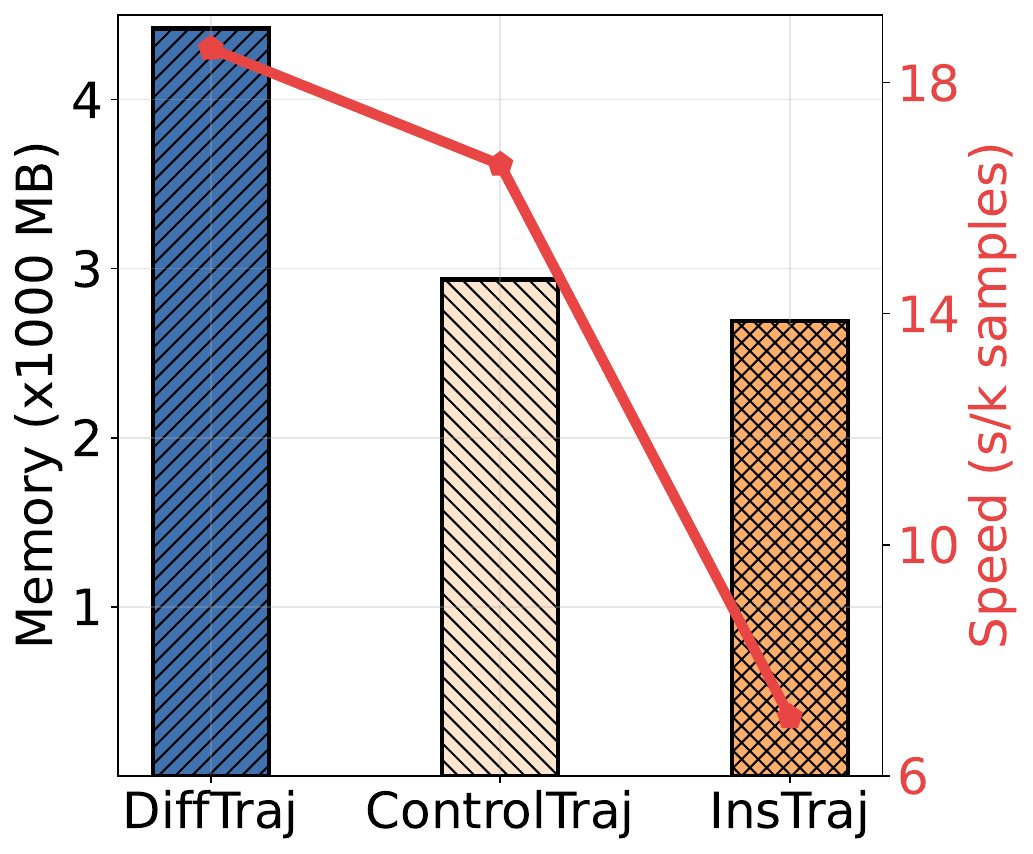}
        \label{fig:sample_efficiency}
    } \hspace{-0.02\linewidth} 
    \subfigure[Impact of patch length.]{
        \includegraphics[width=0.465\linewidth]{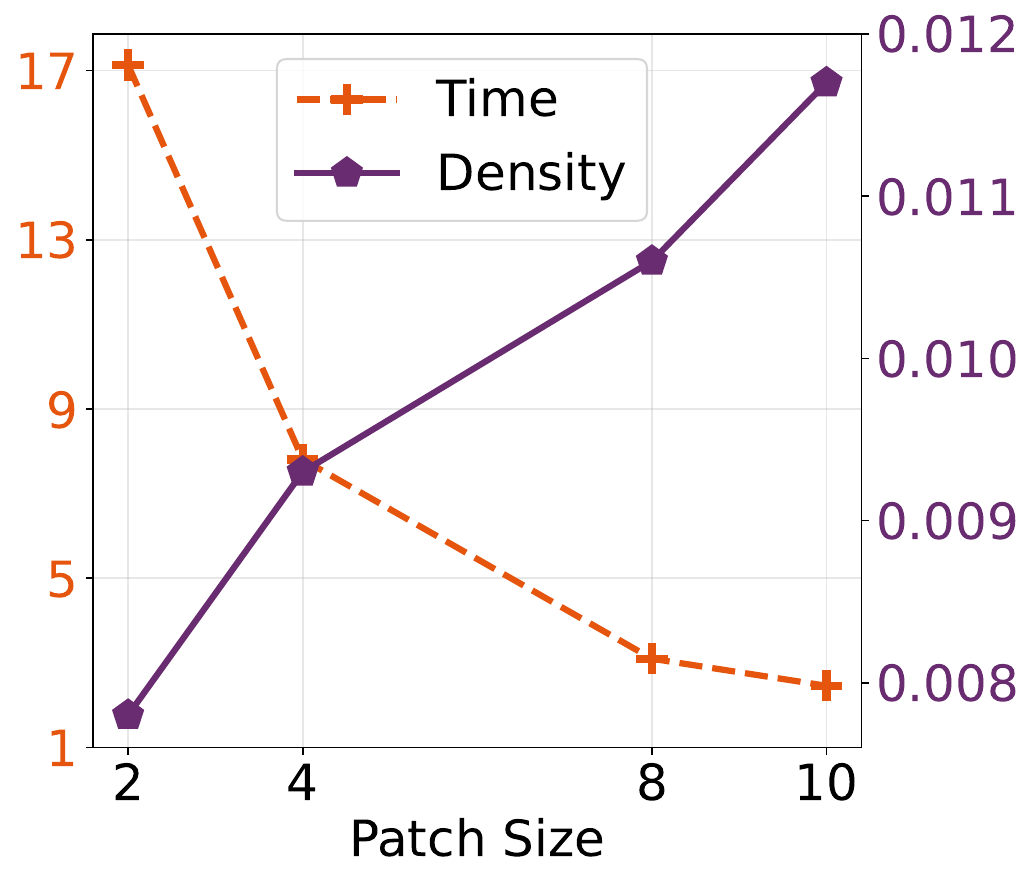}
         \label{fig:patches}
    }
    
    \Description[]{}
   \caption{Performance and efficiency analysis.}
\end{figure}

\subsection{Efficiency and Performance Analysis (RQ3)}
Beyond generative quality and faithfulness, a model's practical utility is critically determined by its computational efficiency. 
We therefore evaluate computational efficiency by measuring the peak GPU memory usage and inference time required to generate 1,000 trajectories with  100 diffusion steps. 
As shown in Figure \ref{fig:sample_efficiency}, \model demonstrates superior efficiency, achieving a $>$2x speedup over both DiffTraj and ControlTraj while maintaining the lowest memory footprint. 
This performance advantage stems from our MT-DiT backbone. 
Unlike the U-Net architectures employed by baselines, which rely on computationally heavy multi-scale convolutions, our Transformer-based design leverages patch-level tokenization and parallel attention mechanisms, significantly reducing the computational overhead per step.

Furthermore, we investigate the impact of a key architectural hyperparameter: the trajectory patch size. 
Figure \ref{fig:patches} reveals a critical trade-off between inference speed and generation fidelity (measured by Density). 
As the patch size increases, the length of the token sequence fed to the Transformer decreases, leading to a significant reduction in inference time. 
Conversely, this speed-up comes at the cost of fidelity; larger patches provide a coarser representation of the trajectory, leading to a lower score. 
This analysis demonstrates that the patch size is a crucial tunable parameter, enabling users to balance computational cost and generative quality based on their specific application needs. 
In summary, the above efficiency results confirm that \model offers compelling advantages in computational performance and a flexible mechanism to navigate the fundamental trade-off between speed and performance.

\subsection{Spatio-Temporal Discovery (RQ4)}
\begin{figure}[t]
    \subfigure[Regular trip.]{
    \includegraphics[width=0.32\linewidth]{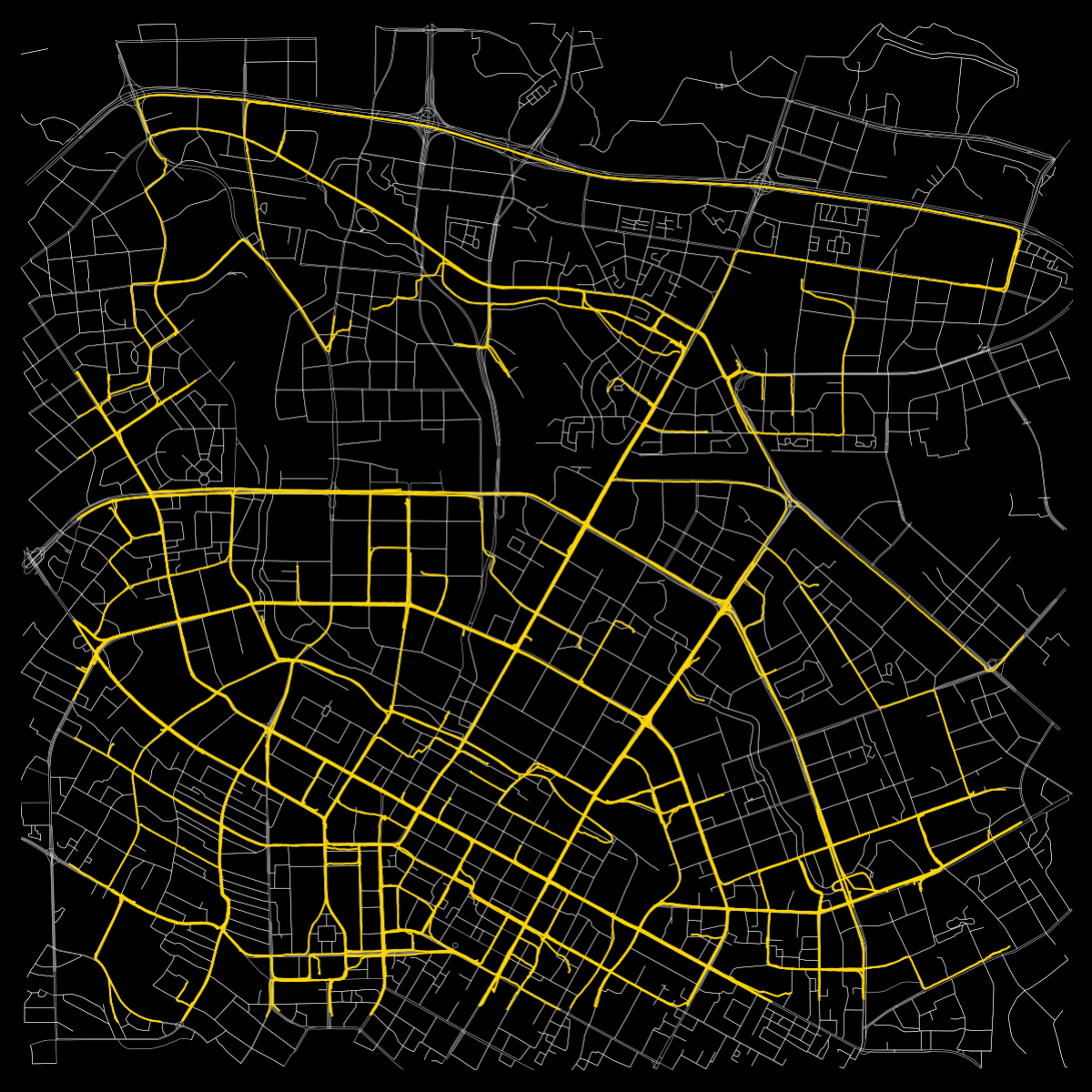}
    } 
    \hspace{-0.03\linewidth} 
    \subfigure[Des. is industrial.]{
    \includegraphics[width=0.32\linewidth]{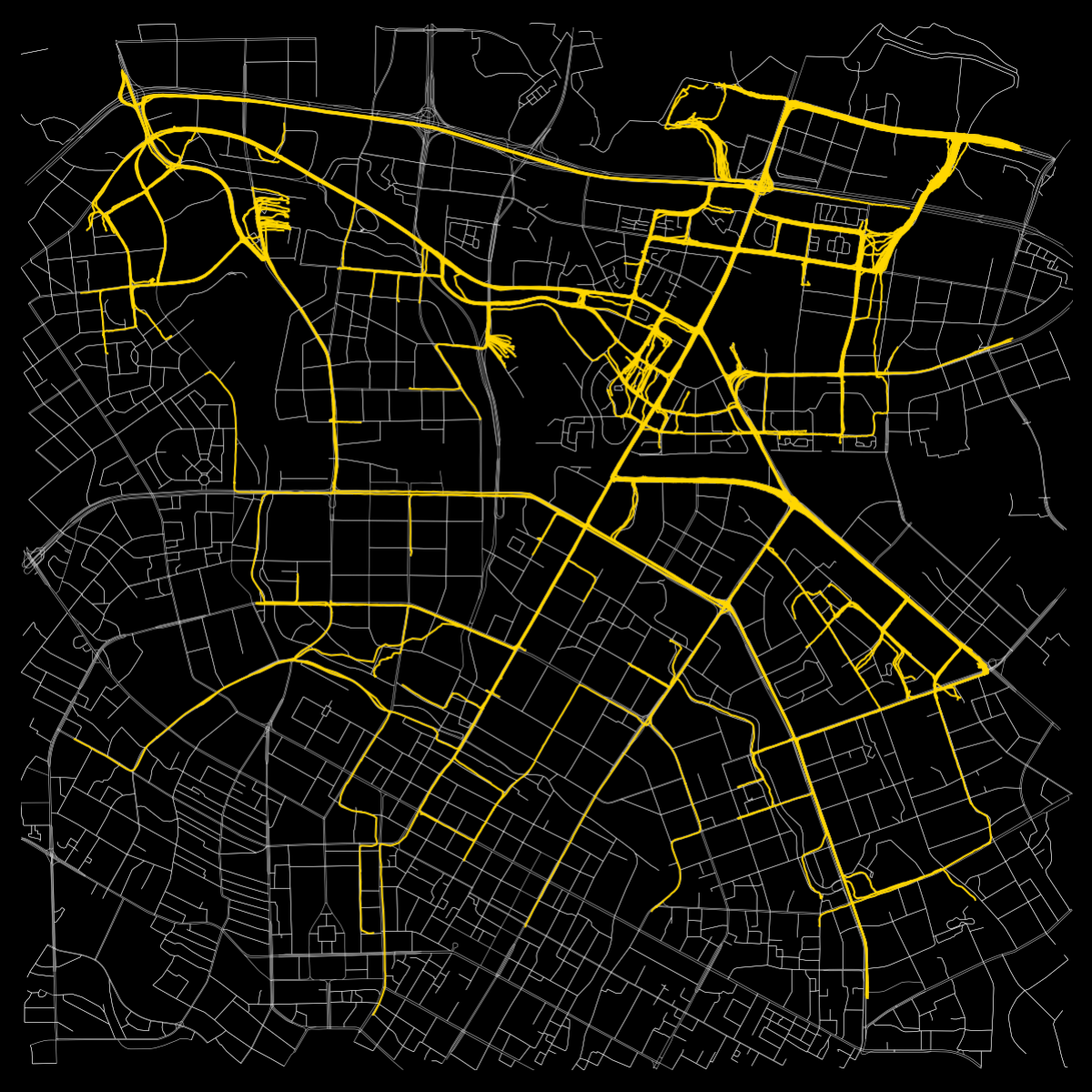}
    }
    \hspace{-0.03\linewidth} 
    \subfigure[Des. is transportation.]{
    \includegraphics[width=0.32\linewidth]{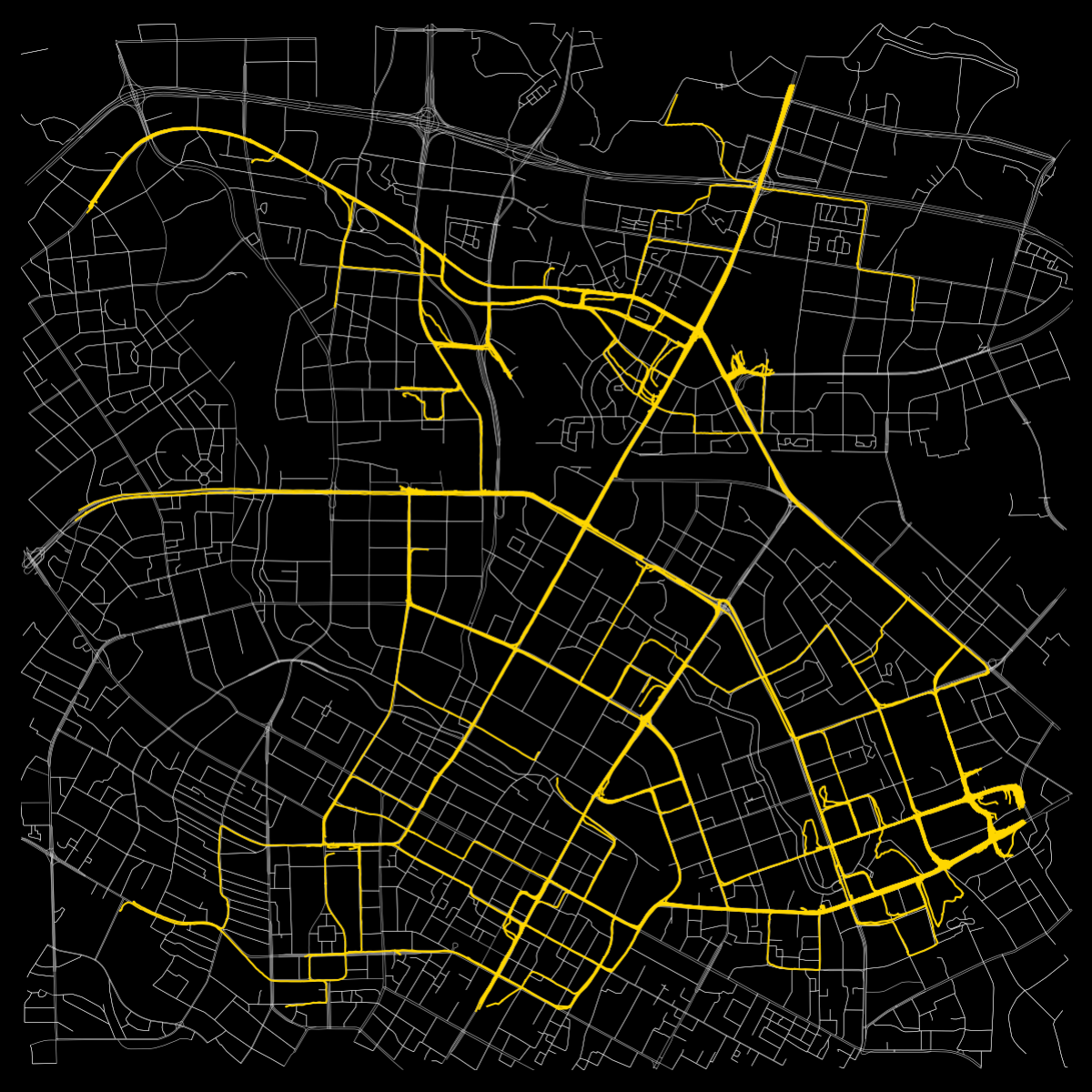}
    }
    \\
    \subfigure[Afternoon periods]{
    \includegraphics[width=0.32\linewidth]{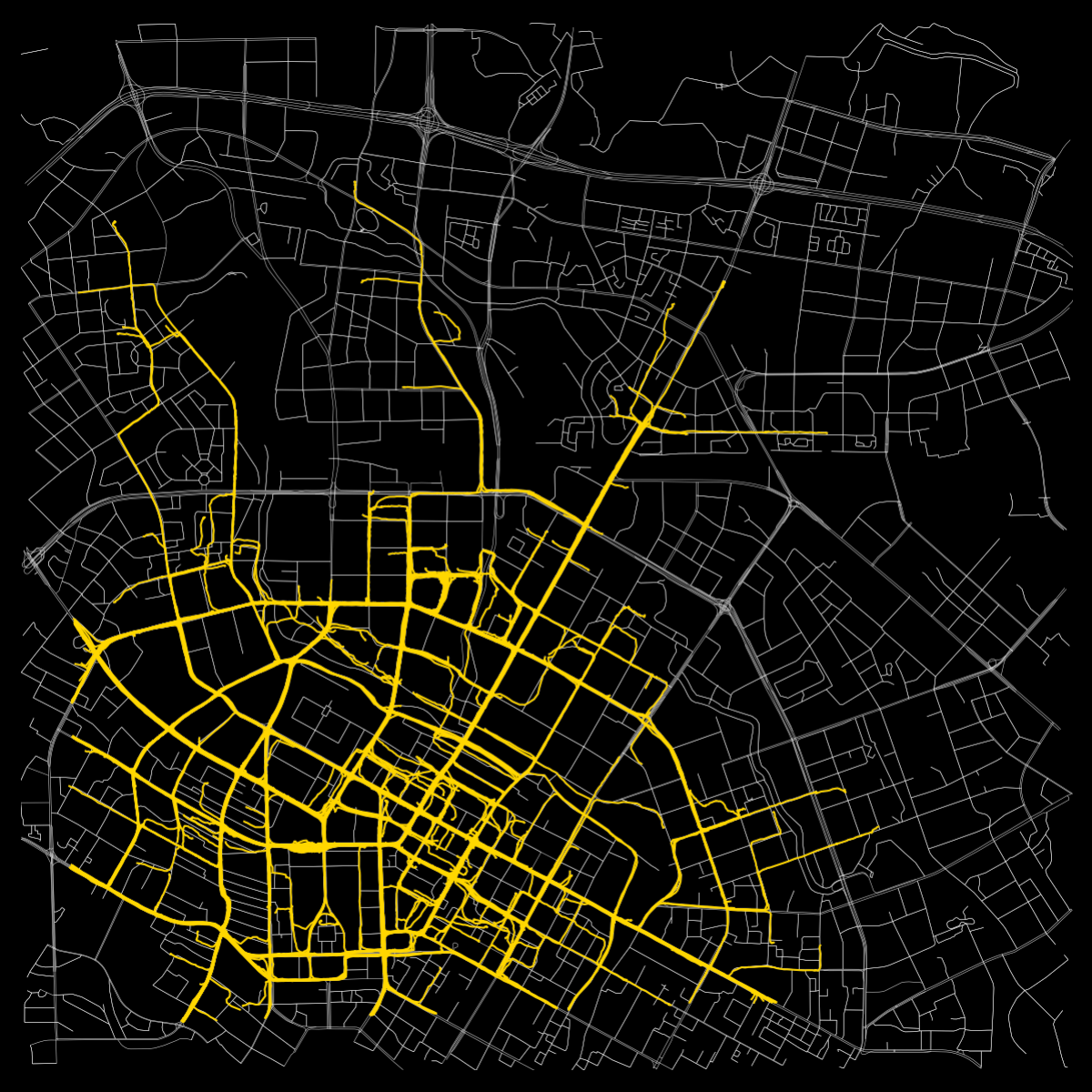}
    }
    \hspace{-0.03\linewidth} 
    \subfigure[Night periods.]{
    \includegraphics[width=0.32\linewidth]{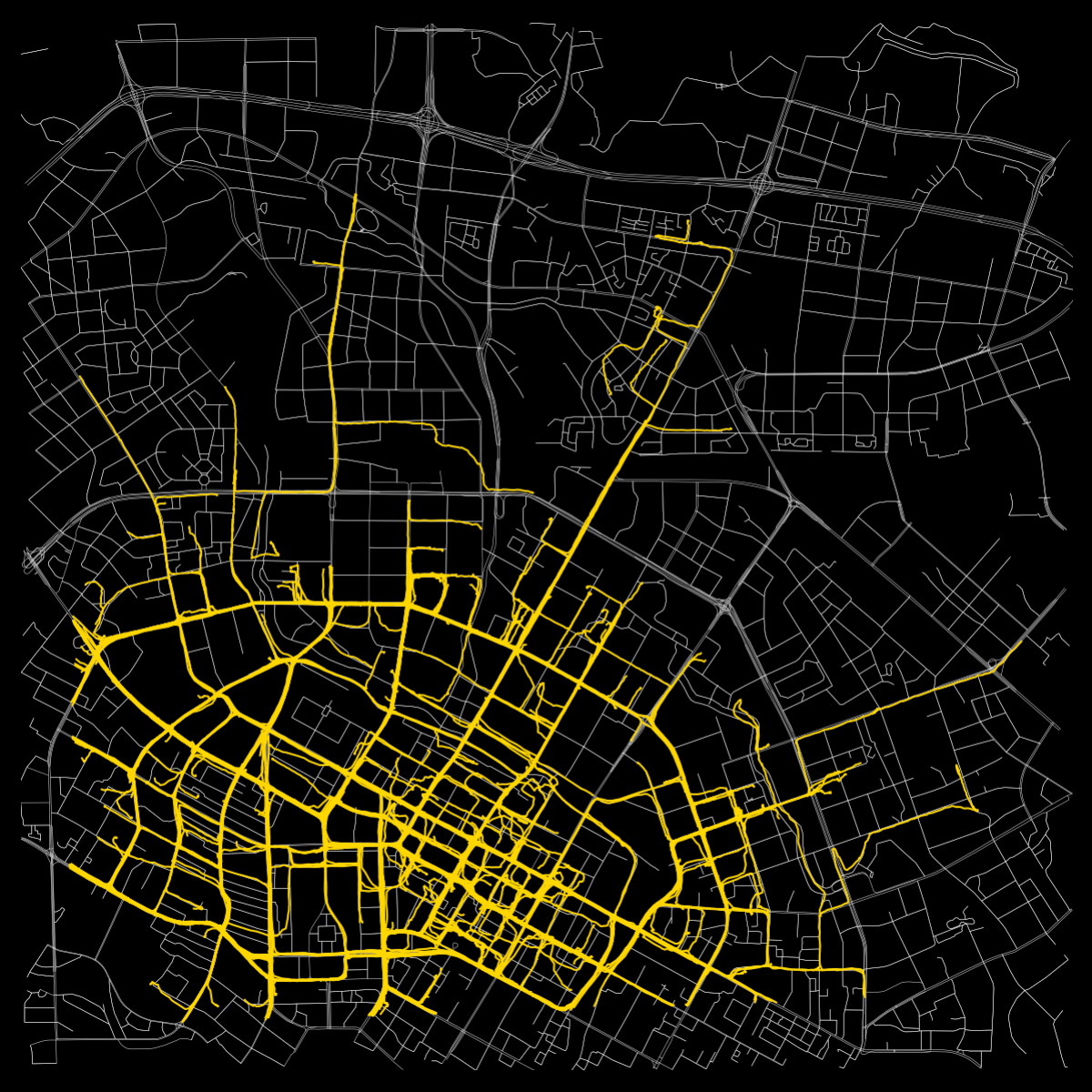}
    }
    \hspace{-0.03\linewidth} 
    \subfigure[Midnight periods.]{
    \includegraphics[width=0.32\linewidth]{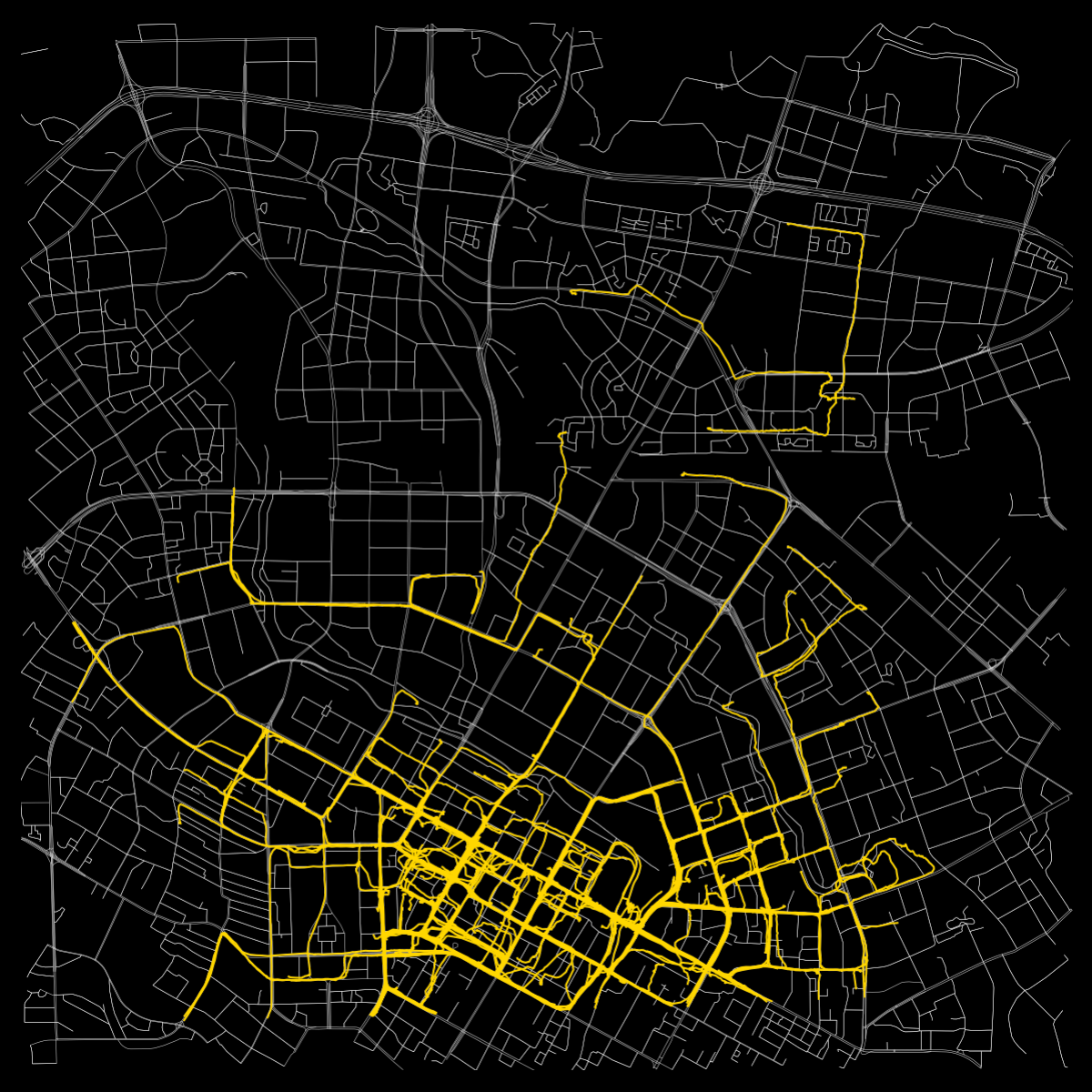}
    } 
      \\
    \subfigure[Short trips.]{
    \includegraphics[width=0.32\linewidth]{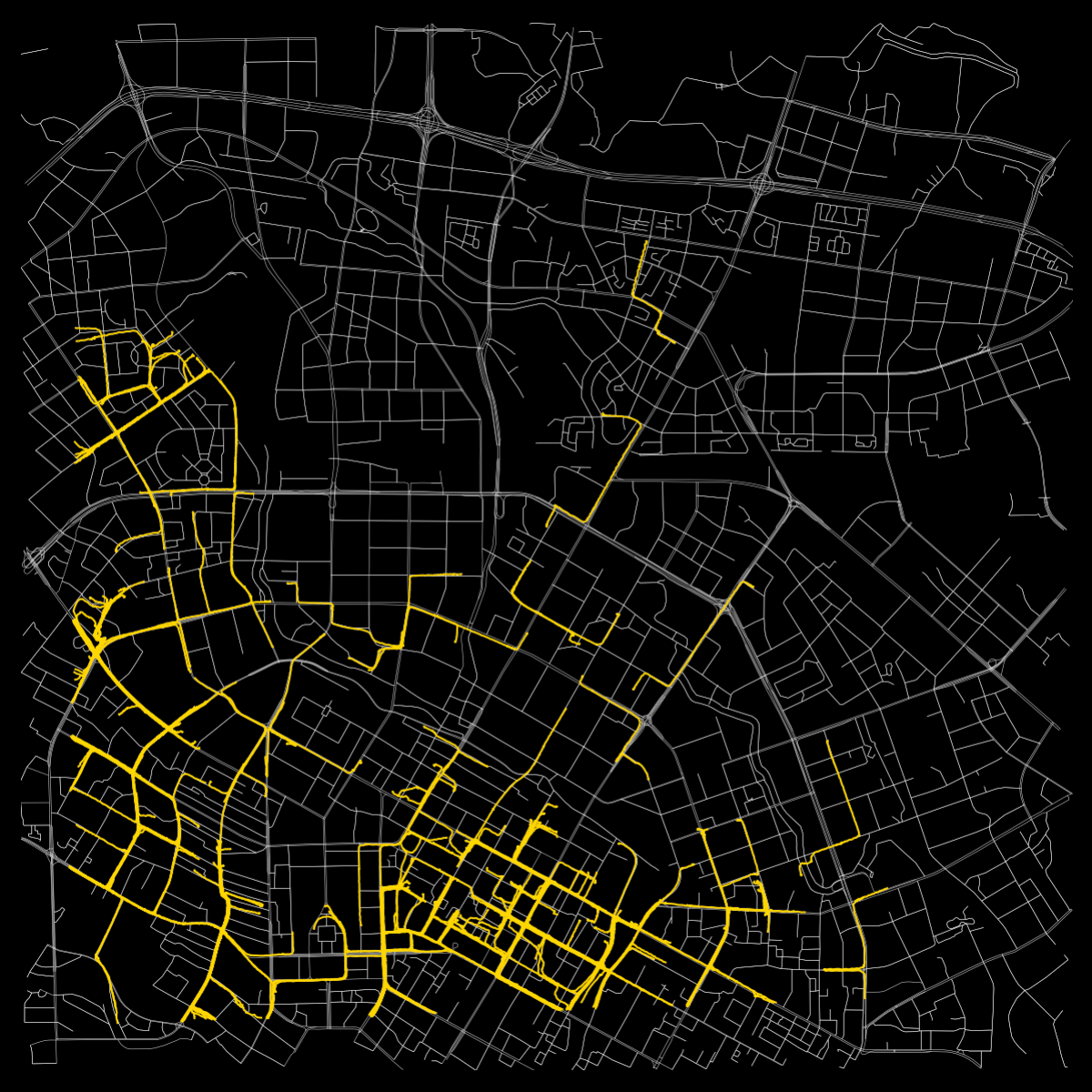}
    } 
    \hspace{-0.03\linewidth} 
    \subfigure[Long trips.]{
    \includegraphics[width=0.32\linewidth]{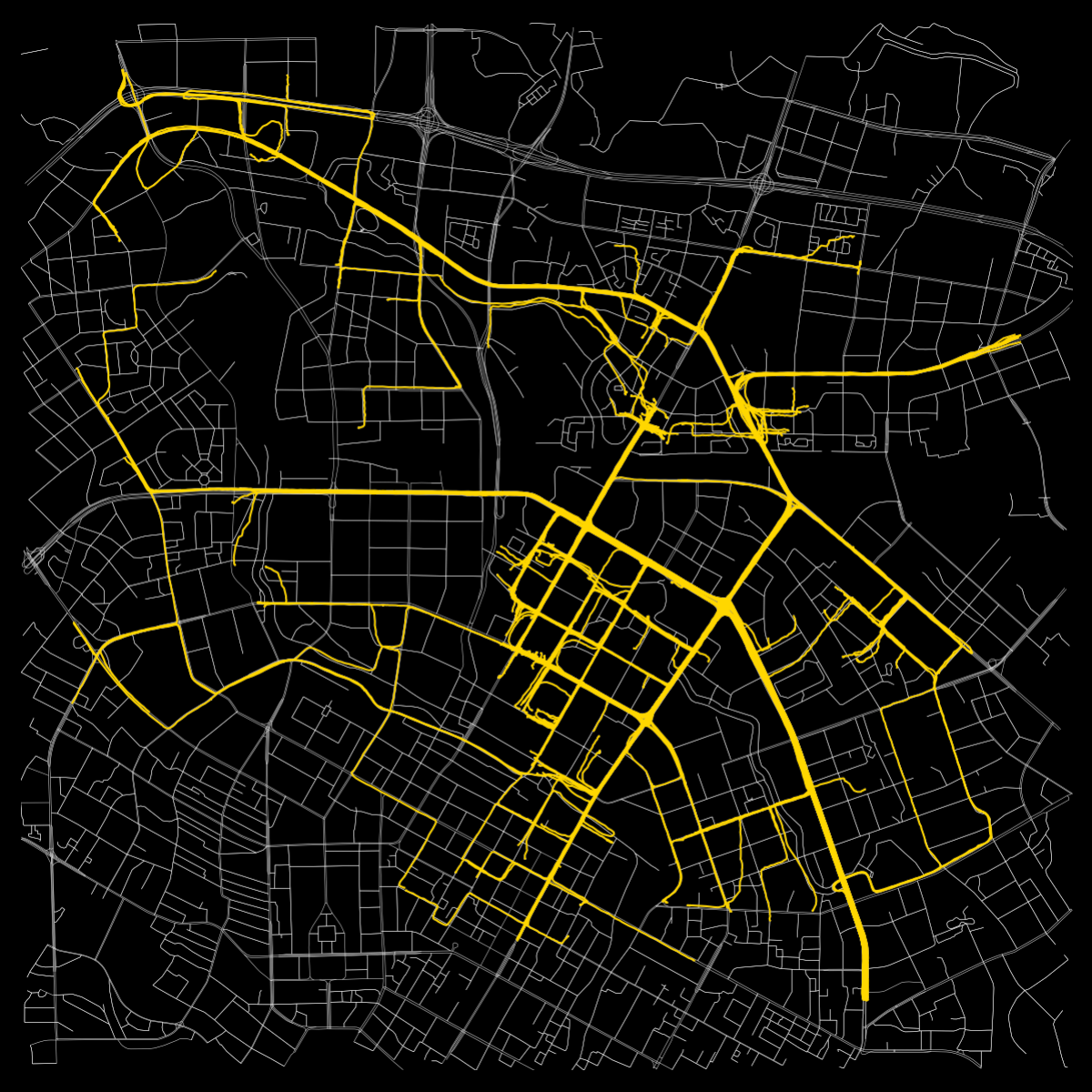}
    }
    \hspace{-0.03\linewidth} 
    \subfigure[Designated areas.]{
    \includegraphics[width=0.32\linewidth]{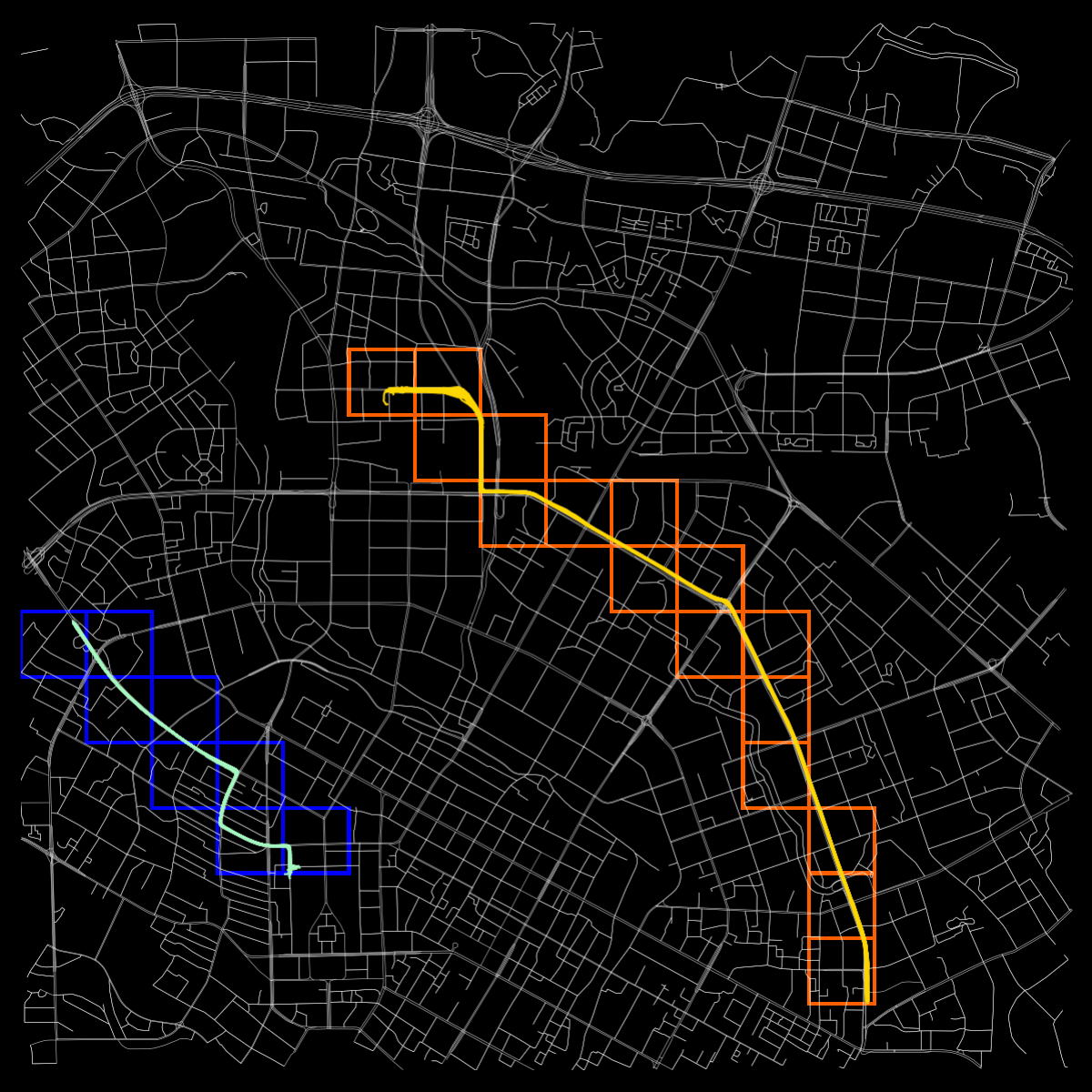}
    }
    \caption{Visualizing spatio-temporal mobility patterns discovered by \model under various instructions guided.}
    \Description[]{}
    \label{fig:st_discovert}
\end{figure}

Finally, we demonstrate \model as a powerful tool for spatio-temporal discovery through multi-level, instruction-based control. 
Figure \ref{fig:st_discovert} illustrates how the model deciphers latent mobility semantics to generate distinct behavioral patterns based on varying instructions.
We analyze these capabilities across three levels.

\noindent \textbf{Purpose-Driven Navigation:} 
The first row (Figure (a)-(c)) illustrates how distinct semantic goals drive trajectory generation. 
Given a common starting context (e.g., residential area), we observe that instructions specifying different purposes trigger significantly divergent path behaviors. 
Specifically, general or unspecified commands result in diffuse movement (a) that covers the entire urban area.
When we adjust the semantic destination to industrial areas (b) or transportation areas (c), we observe entirely different distribution patterns.
It activates the model's learned knowledge of the city's functional layout, causing trajectories to naturally navigate towards the corresponding functional zones. 
This confirms that \model has learned to map high-level semantic purposes to specific, geographically-aware mobility patterns.

\noindent \textbf{Temporal Pattern Discovery:}
The second row (Figure (d)-(f)) highlights the \model's ability to interpret and replicate temporal semantics. 
Rather than simply treating ``Afternoon or Midnight'' as text labels, \model captures the underlying traffic dynamics associated with these time periods.
As shown in these figures, trajectories generated by instructions at midnight (departure time at midnight) exhibit sparse traffic patterns concentrated on major thoroughfares, whereas those generated by instructions in the afternoon (departure time in the afternoon) exhibit dense, widespread traffic activity across local streets.
This indicates that \model has successfully internalized the city's latent temporal rhythms, enabling it to reproduce complex, time-dependent traffic distributions solely from semantic descriptors.

\noindent \textbf{Fine-Grained Attribute and Constraint Control:} 
The bottom row (Figure (g)-(i)) demonstrates the model's conceptual understanding of spatial attributes. 
Instead of relying on explicit distance thresholds, \model can automatically interpret distance-related descriptions in instructions and translate these relative concepts into concrete geometric behaviors.
For example, it can generate distinct distributions of short paths (g) and long paths (h) based on varying distances.
Most impressively, Figure \ref{fig:st_discovert}(i) shows that \model can handle novel, dynamic spatial constraints by generating a plausible route between arbitrarily ``Designated areas''. 
This showcases a capability beyond mere pattern replication, bordering on genuine spatial reasoning and planning.
This level of fine-grained control is where \model truly excels, offering a flexible interface for manipulating nuanced travel characteristics that are impossible to specify in traditional trajectory-generation systems.
To rigorously validate this, we conducted a quantitative evaluation for the origin-destination control in \textbf{Appendix \ref{app:odanalysis}}.

In summary, these case studies affirm that \model is more than a generative model; it is an effective simulation engine for urban discovery, enabling users to explore a wide variety of mobility scenarios simply by describing them in natural language.
\vspace{-3mm}
\section{Related work}\label{sec:related}

\noindent \textbf{Generative Models for Trajectory Data.}
Generative modeling of GPS trajectories has evolved significantly with the advent of deep learning \cite{luca2021survey}. 
While early statistical models laid the groundwork, they often struggled to capture the complex, non-linear dynamics of human mobility. The field saw a substantial leap with Recurrent Neural Networks, such as in DeepMove \cite{feng2018deepmove}, which effectively modeled the sequential nature of trajectory data. 
Subsequently, more powerful frameworks like Generative Adversarial Networks (GANs) and Variational Autoencoders (VAEs) were adapted for this task. Models such as TrajGAN \cite{rao2020lstm, liu2018trajgans} and Dp-TrajGAN focused on enhancing realism and incorporating a degree of structured control, for instance, by conditioning on destinations.
More recently, denoising diffusion probabilistic models have emerged as the new state-of-the-art. DiffTraj \cite{zhu2023difftraj} pioneered their application, demonstrating an unprecedented ability to generate high-fidelity and diverse trajectories. 
Building on this, ControlTraj \cite{zhu2024controltraj} represents the current frontier in controllable generation by conditioning the diffusion process on road network topology. 
This allows the model to produce geographically plausible paths that adhere to physical infrastructure.
Further research has explored the use of diffusion models to accomplish more specific scenarios, such as long-term trajectory prediction and the generation of trajectories with certain social or behavioral attributes, demonstrating the potential of this generative paradigm \cite{wei2024diff,cao2025holistic,zhang2025noise,long2025one}.
While these deep learning models have advanced in trajectory generation, they often lack an intuitive mechanism for incorporating high-level semantic instructions from human users. 
Their conditioning mechanisms are typically limited to structured inputs (e.g., start/end locations, time) and do not inherently support the rich, unstructured nature of natural language commands, which is a key focus of our framework.

\noindent \textbf{Conditioned Diffusion Models.}
A significant line of research has extended diffusion models to incorporate conditioning, enabling fine-grained control over the generative process \cite{oppenlaender2022creativity}. While various conditioning mechanisms exist, the most transformative has been the use of natural language with specifically in Text-to-Image synthesis \cite{cao2025controllable, li2019controllable}. 
This revolution was catalyzed by the development of powerful multi-modal models like CLIP \cite{radford2021learning}, which learned to align text and image representations in a shared semantic space. 
Landmark diffusion models with this paradigm, such as GLIDE \cite{nichol2021glide}, DALL-E-2, and Imagen \cite{saharia2022photorealistic}, leveraged this capability to generate highly detailed and contextually-aware images from complex text prompts.
Subsequently, Latent Diffusion Models (Stable Diffusion) \cite{rombach2022high} significantly improved computational efficiency by performing diffusion in a compressed latent space. 
More recently, to achieve precise structural control beyond text prompts, frameworks like ControlNet \cite{zhang2023adding} and T2I-Adapter \cite{mou2024t2i} were introduced to support additional spatial conditions (e.g., canny edges, depth maps).
Furthermore, the generative backbone has evolved from traditional U-Nets to Diffusion Transformers (DiT) \cite{peebles2023scalable}, demonstrating the scalability of Transformer architectures in diffusion processes.
This, combined with training techniques like Classifier-Free Guidance \cite{Ho2022ClassifierFreeDG}, provided a robust way to amplify the influence of the conditioning signal, ensuring faithfulness to the prompt.
This powerful ``Text-to-X'' paradigm, which pairs a strong language encoder with a deep fusion mechanism, has proven highly versatile and has been successfully extended to other domains, such as video generation \cite{singer2022make} and speech synthesis \cite{liu2023diffvoice}.
However, despite its transformative impact, its application to the synthesis of spatio-temporal trajectory data remains largely unexplored.  
Our work, \model, is the first to systematically bridge this gap by adapting and tailoring this successful language-conditioned generative paradigm specifically for the unique characteristics of human mobility data.

\section{Conclusion and Future Work}\label{sec:conclusion}

In this paper, we introduce \model, a novel framework that generates high-fidelity trajectories directly from natural language instructions. By synergizing an LLM-based semantic encoder with the MT-DiT architecture, \model effectively bridges the gap between unstructured intentions and continuous spatio-temporal dynamics, enabling fine-grained semantic control. 
Despite these advancements, limitations exist: First, our evaluation is currently restricted to taxi trajectory datasets. While these serve as robust benchmarks, they may not fully capture the distinct behavioral patterns of other mobility modes.  Second, although our experiments demonstrate effective transferability via fine-tuning, achieving zero-shot generalization across cities remains a challenge due to the significant topological heterogeneity.
In the future, we plan to explore incorporating broader mobility data and explicit topological priors to decouple generation from specific city layouts and travel modes, thereby enhancing the model's generative capabilities in dynamic environments.


\bibliographystyle{ACM-Reference-Format}
\bibliography{ref}

@String{Computing = "Computing" }

@String{Computer = "{IEEE} Computer" }

@String{Academic = "Academic Press" }

@String{Springer = "Springer-Verlag" }

@article{ddpm,
  title={Denoising diffusion probabilistic models},
  author={Ho, Jonathan and Jain, Ajay and Abbeel, Pieter},
  journal={Advances in neural information processing systems},
  volume={33},
  pages={6840--6851},
  year={2020}
}

@article{zhang2022dp,
  title={DP-TrajGAN: A privacy-aware trajectory generation model with differential privacy},
  author={Zhang, Jing and Huang, Qihan and Huang, Yirui and Ding, Qian and Tsai, Pei-Wei},
  journal={Future Generation Computer Systems},
  year={2022},
  publisher={Elsevier}
}

@article{luca2021survey,
  title={A survey on deep learning for human mobility},
  author={Luca, Massimiliano and Barlacchi, Gianni and Lepri, Bruno and Pappalardo, Luca},
  journal={ACM Computing Surveys (CSUR)},
  volume={55},
  number={1},
  pages={1--44},
  year={2021},
  publisher={ACM New York, NY}
}

@article{diffusionsuvery,
  title={Diffusion models: A comprehensive survey of methods and applications},
  author={Yang, Ling and Zhang, Zhilong and Song, Yang and Hong, Shenda and Xu, Runsheng and Zhao, Yue and Zhang, Wentao and Cui, Bin and Yang, Ming-Hsuan},
  journal={ACM computing surveys},
  volume={56},
  number={4},
  pages={1--39},
  year={2023},
  publisher={ACM New York, NY, USA}
}

@inproceedings{liu2018trajgans,
  title={trajGANs: Using generative adversarial networks for geo-privacy protection of trajectory data (Vision paper)},
  author={Liu, Xia and Chen, Hanzhou and Andris, Clio},
  booktitle={Location privacy and security workshop},
  pages={1--7},
  year={2018}
}

@inproceedings{nichol2021glide,
  title={GLIDE: Towards Photorealistic Image Generation and Editing with Text-Guided Diffusion Models},
  author={Nichol, Alexander Quinn and Dhariwal, Prafulla and Ramesh, Aditya and Shyam, Pranav and Mishkin, Pamela and Mcgrew, Bob and Sutskever, Ilya and Chen, Mark},
  booktitle={International Conference on Machine Learning},
  pages={16784--16804},
  year={2022},
  organization={PMLR}
}

@inproceedings{rombach2022high,
  title={High-resolution image synthesis with latent diffusion models},
  author={Rombach, Robin and Blattmann, Andreas and Lorenz, Dominik and Esser, Patrick and Ommer, Bj{\"o}rn},
  booktitle={Proceedings of the IEEE/CVF conference on computer vision and pattern recognition},
  pages={10684--10695},
  year={2022}
}

@article{Ho2022ClassifierFreeDG,
  title={Classifier-free diffusion guidance},
  author={Ho, Jonathan and Salimans, Tim},
  journal={arXiv preprint arXiv:2207.12598},
  year={2022}
}

@inproceedings{kongdiffwave,
    title={DiffWave: A Versatile Diffusion Model for Audio Synthesis},
    author={Kong, Zhifeng and Ping, Wei and Huang, Jiaji and Zhao, Kexin and Catanzaro, Bryan},
    year={2020},
    booktitle={International Conference on Learning Representations}
}

@article{chen2024deep,
  title={Deep learning for trajectory data management and mining: A survey and beyond},
  author={Chen, Wei and Liang, Yuxuan and Zhu, Yuanshao and Chang, Yanchuan and Luo, Kang and Wen, Haomin and Li, Lei and Yu, Yanwei and Wen, Qingsong and Chen, Chao and others},
  journal={arXiv preprint arXiv:2403.14151},
  year={2024}
}

@inproceedings{zhu2023difftraj,
    title={DiffTraj: Generating {GPS} Trajectory with Diffusion Probabilistic Model},
    author={Yuanshao Zhu and Yongchao Ye and Shiyao Zhang and Xiangyu Zhao and James Jianqiao Yu},
    booktitle={Thirty-seventh Conference on Neural Information Processing Systems},
    year={2023}
}

@InProceedings{rao2020lstm,
  title =	{{LSTM-TrajGAN}: A Deep Learning Approach to Trajectory Privacy Protection},
  author={Rao, Jinmeng and Gao, Song and Kang, Yuhao and Huang, Qunying},
  booktitle =	{11th International Conference on Geographic Information Science (GIScience 2021)},
  year={2020}
}

@inproceedings{radford2021learning,
  title={Learning transferable visual models from natural language supervision},
  author={Radford, Alec and Kim, Jong Wook and Hallacy, Chris and Ramesh, Aditya and Goh, Gabriel and Agarwal, Sandhini and Sastry, Girish and Askell, Amanda and Mishkin, Pamela and Clark, Jack and others},
  booktitle={International conference on machine learning},
  pages={8748--8763},
  year={2021},
  organization={PMLR}
}

@inproceedings{guo2018learning,
  title={Learning to route with sparse trajectory sets},
  author={Guo, Chenjuan and Yang, Bin and Hu, Jilin and Jensen, Christian},
  booktitle={2018 IEEE 34th International Conference on Data Engineering (ICDE)},
  pages={1073--1084},
  year={2018},
  organization={IEEE}
}

@inproceedings{devlin2019bert,
  title={Bert: Pre-training of deep bidirectional transformers for language understanding},
  author={Devlin, Jacob and Chang, Ming-Wei and Lee, Kenton and Toutanova, Kristina},
  booktitle={Proceedings of the 2019 conference of the North American chapter of the association for computational linguistics: human language technologies,},
  pages={4171--4186},
  year={2019}
}

@inproceedings{feng2018deepmove,
  title={Deepmove: Predicting human mobility with attentional recurrent networks},
  author={Feng, Jie and Li, Yong and Zhang, Chao and Sun, Funing and Meng, Fanchao and Guo, Ang and Jin, Depeng},
  booktitle={Proceedings of the 2018 world wide web conference},
  pages={1459--1468},
  year={2018}
}

@inproceedings{zhu2024controltraj,
  title={Controltraj: Controllable trajectory generation with topology-constrained diffusion model},
  author={Zhu, Yuanshao and Yu, James Jianqiao and Zhao, Xiangyu and Liu, Qidong and Ye, Yongchao and Chen, Wei and Zhang, Zijian and Wei, Xuetao and Liang, Yuxuan},
  booktitle={Proceedings of the 30th ACM SIGKDD Conference on Knowledge Discovery and Data Mining},
  pages={4676--4687},
  year={2024}
}

@article{xu2024mm,
  title={MM-Path: Multi-modal, Multi-granularity Path Representation Learning--Extended Version},
  author={Xu, Ronghui and Cheng, Hanyin and Guo, Chenjuan and Gao, Hongfan and Hu, Jilin and Yang, Sean Bin and Yang, Bin},
  journal={arXiv preprint arXiv:2411.18428},
  year={2024}
}

@article{guo2020context,
  title={Context-aware, preference-based vehicle routing},
  author={Guo, Chenjuan and Yang, Bin and Hu, Jilin and Jensen, Christian S and Chen, Lu},
  journal={The VLDB Journal},
  volume={29},
  pages={1149--1170},
  year={2020},
  publisher={Springer}
}

@inproceedings{esser2024scaling,
  title={Scaling rectified flow transformers for high-resolution image synthesis},
  author={Esser, Patrick and Kulal, Sumith and Blattmann, Andreas and Entezari, Rahim and M{\"u}ller, Jonas and Saini, Harry and Levi, Yam and Lorenz, Dominik and Sauer, Axel and Boesel, Frederic and others},
  booktitle={Forty-first international conference on machine learning},
  year={2024}
}

@inproceedings{gao2021simcse,
   title={{SimCSE}: Simple Contrastive Learning of Sentence Embeddings},
   author={Gao, Tianyu and Yao, Xingcheng and Chen, Danqi},
   booktitle={Empirical Methods in Natural Language Processing (EMNLP)},
   year={2021}
}

@article{saharia2022photorealistic,
  title={Photorealistic text-to-image diffusion models with deep language understanding},
  author={Saharia, Chitwan and Chan, William and Saxena, Saurabh and Li, Lala and Whang, Jay and Denton, Emily L and Ghasemipour, Kamyar and Gontijo Lopes, Raphael and Karagol Ayan, Burcu and Salimans, Tim and others},
  journal={Advances in neural information processing systems},
  volume={35},
  pages={36479--36494},
  year={2022}
}

@inproceedings{singer2022make,
  title={Make-A-Video: Text-to-Video Generation without Text-Video Data},
  author={Singer, Uriel and Polyak, Adam and Hayes, Thomas and Yin, Xi and An, Jie and Zhang, Songyang and Hu, Qiyuan and Yang, Harry and Ashual, Oron and Gafni, Oran and others},
  booktitle={The Eleventh International Conference on Learning Representations}
}

@article{wei2024diff,
  title={Diff-rntraj: A structure-aware diffusion model for road network-constrained trajectory generation},
  author={Wei, Tonglong and Lin, Youfang and Guo, Shengnan and Lin, Yan and Huang, Yiheng and Xiang, Chenyang and Bai, Yuqing and Wan, Huaiyu},
  journal={IEEE Transactions on Knowledge and Data Engineering},
  year={2024},
  publisher={IEEE}
}

@article{cao2025holistic,
  title={Holistic Semantic Representation for Navigational Trajectory Generation},
  author={Cao, Ji and Zheng, Tongya and Guo, Qinghong and Wang, Yu and Dai, Junshu and Liu, Shunyu and Yang, Jie and Song, Jie and Song, Mingli},
  journal={arXiv preprint arXiv:2501.02737},
  year={2025}
}

@inproceedings{zhang2025noise,
  title={Noise Matters: Diffusion Model-based Urban Mobility Generation with Collaborative Noise Priors},
  author={Zhang, Yuheng and Yuan, Yuan and Ding, Jingtao and Yuan, Jian and Li, Yong},
  booktitle={Proceedings of the ACM on Web Conference 2025},
  pages={5352--5363},
  year={2025}
}

@article{chen2021trajvae,
  title={Trajvae: A variational autoencoder model for trajectory generation},
  author={Chen, Xinyu and Xu, Jiajie and Zhou, Rui and Chen, Wei and Fang, Junhua and Liu, Chengfei},
  journal={Neurocomputing},
  volume={428},
  pages={332--339},
  year={2021},
  publisher={Elsevier}
}

@article{yang2024survey,
  title={A survey on diffusion models for time series and spatio-temporal data},
  author={Yang, Yiyuan and Jin, Ming and Wen, Haomin and Zhang, Chaoli and Liang, Yuxuan and Ma, Lintao and Wang, Yi and Liu, Chenghao and Yang, Bin and Xu, Zenglin and others},
  journal={arXiv preprint arXiv:2404.18886},
  year={2024}
}

@article{long2025one,
  title={One Fits All: General Mobility Trajectory Modeling via Masked Conditional Diffusion},
  author={Long, Qingyue and Rong, Can and Wang, Huandong and Li, Yong},
  journal={arXiv preprint arXiv:2501.13347},
  year={2025}
}

@article{li2019controllable,
  title={Controllable text-to-image generation},
  author={Li, Bowen and Qi, Xiaojuan and Lukasiewicz, Thomas and Torr, Philip},
  journal={Advances in neural information processing systems},
  volume={32},
  year={2019}
}

@inproceedings{oppenlaender2022creativity,
  title={The creativity of text-to-image generation},
  author={Oppenlaender, Jonas},
  booktitle={Proceedings of the 25th international academic mindtrek conference},
  pages={192--202},
  year={2022}
}

@inproceedings{mou2024t2i,
  title={T2i-adapter: Learning adapters to dig out more controllable ability for text-to-image diffusion models},
  author={Mou, Chong and Wang, Xintao and Xie, Liangbin and Wu, Yanze and Zhang, Jian and Qi, Zhongang and Shan, Ying},
  booktitle={Proceedings of the AAAI conference on artificial intelligence},
  volume={38},
  number={5},
  pages={4296--4304},
  year={2024}
}

@inproceedings{zhang2023adding,
  title={Adding conditional control to text-to-image diffusion models},
  author={Zhang, Lvmin and Rao, Anyi and Agrawala, Maneesh},
  booktitle={Proceedings of the IEEE/CVF international conference on computer vision},
  pages={3836--3847},
  year={2023}
}

@inproceedings{liu2023diffvoice,
  title={Diffvoice: Text-to-speech with latent diffusion},
  author={Liu, Zhijun and Guo, Yiwei and Yu, Kai},
  booktitle={IEEE International Conference on Acoustics, Speech and Signal Processing (ICASSP)},
  year={2023},
  organization={IEEE}
}

@inproceedings{peebles2023scalable,
  title={Scalable diffusion models with transformers},
  author={Peebles, William and Xie, Saining},
  booktitle={Proceedings of the IEEE/CVF international conference on computer vision},
  pages={4195--4205},
  year={2023}
}

@article{peng2025diffusion,
  title={Diffusion models for intelligent transportation systems: A survey},
  author={Peng, Mingxing and Chen, Kehua and Guo, Xusen and Zhang, Qiming and Zhong, Hui and Zhu, Meixin and Yang, Hai},
  journal={IEEE Transactions on Intelligent Transportation Systems},
  year={2025},
  publisher={IEEE}
}

@article{touvron2023llama,
  title={Llama: Open and efficient foundation language models},
  author={Touvron, Hugo and Lavril, Thibaut and Izacard, Gautier and Martinet, Xavier and Lachaux, Marie-Anne and Lacroix, Timoth{\'e}e and Rozi{\`e}re, Baptiste and Goyal, Naman and Hambro, Eric and Azhar, Faisal and others},
  journal={arXiv preprint arXiv:2302.13971},
  year={2023}
}

@article{achiam2023gpt,
  title={Gpt-4 technical report},
  author={Achiam, Josh and Adler, Steven and Agarwal, Sandhini and Ahmad, Lama and Akkaya, Ilge and Aleman, Florencia Leoni and Almeida, Diogo and Altenschmidt, Janko and Altman, Sam and Anadkat, Shyamal and others},
  journal={arXiv preprint arXiv:2303.08774},
  year={2023}
}

@article{llm2vec,
  title={Llm2vec: Large language models are secretly powerful text encoders},
  author={BehnamGhader, Parishad and Adlakha, Vaibhav and Mosbach, Marius and Bahdanau, Dzmitry and Chapados, Nicolas and Reddy, Siva},
  journal={arXiv preprint arXiv:2404.05961},
  year={2024}
}

@inproceedings{cao2021generating,
  title={Generating mobility trajectories with retained data utility},
  author={Cao, Chu and Li, Mo},
  booktitle={Proceedings of the 27th ACM SIGKDD conference on knowledge discovery \& data mining},
  pages={2610--2620},
  year={2021}
}

@article{cao2025controllable,
  title={Controllable generation with text-to-image diffusion models: A survey},
  author={Cao, Pu and Zhou, Feng and Song, Qing and Yang, Lu},
  journal={IEEE Transactions on Pattern Analysis and Machine Intelligence},
  year={2025},
  publisher={IEEE}
}

\clearpage
\begin{appendix}
\section{Instruction Construction Pipeline}\label{app:data_construction}
This section will describe the construction process of the human intent instructions used in this paper.

\subsection{Regional Function Profiling} 
To construct a semantic map of the urban environment, the area of each dataset was partitioned into a grid of 50m $\times$ 50m cells.
We extracted Point-of-Interest (POI) data from an OpenStreetMap (OSM) snapshot dated December 2024.
These raw POIs were mapped to average $K=18$ distinct functional categories to capture the diverse urban landscape.
A summary of this mapping schema is provided in Table \ref{tab:poi_mapping_appendix}.
For instance, tags such as apartment, residential are mapped to \textit{Residential}, while mall, supermarket, restaurant are mapped to \textit{Commercial}.
We calculate the functional vector $\boldsymbol{f}_{i,j}$ for each grid cell by normalizing the frequency of these POI categories.

\begin{table}[h]
\centering
\caption{Mapping schema from OSM tags to functional categories.}
\vspace{-3mm}
\label{tab:poi_mapping_appendix}
\resizebox{0.98\linewidth}{!}{
\begin{tabular}{ll}
\toprule
\textbf{Functional Category} & \textbf{Example OSM Tags} \\
\midrule
Residential & building=residential, landuse=residential \\
Commercial & shop=*, building=retail \\
Food & amenity=restaurant, amenity=cafe \\
Transportation & highway=bus stop, railway=station \\
Education & building=school, amenity=university \\
Recreational & leisure=stadium, leisure=playground \\
Green space & leisure=park, leisure=garden \\
... & ... \\
\bottomrule
\end{tabular}
}
\end{table}

\subsection{Instruction Generation} 
To bridge the gap between structured trajectory data and natural language, we implement a generation pipeline.
We create templates for different travel intentions, including general travel, going to work, going shopping, etc. For each set of templates, we use the GPT-4 model from OpenAI to generate more synonyms and produce diverse, human-like descriptions. The prompt design is shown below:
\begin{tcolorbox}[title=System Prompt for Instruction Refinement, colback=gray!5!white,colframe=gray!75!black,fonttitle=\bfseries]
\textbf{Role:} You are an expert in urban mobility behavior analysis. \\
\textbf{Task:} Transform the provided structured mobility attributes into a natural language description of a trip intention. \\ 
\textbf{Input Attributes:} [Origin: Residential], [Destination: Green Space], [Distance: xx km], [Time: xx timeslot] \\
\textbf{Requirements:} \\
\textbf{Diversity:} Use varied sentence structures, voice, and different synonyms for connection. \\
\textbf{Naturalness:} The tone should simulate a casual human description or a navigation command. \\
\textbf{Accuracy:} Strictly adhere to the provided attributes without hallucinating new details. 
\end{tcolorbox}
Using this pipeline, we constructed a hierarchical instruction dataset ranging from abstract to concrete levels. Examples include:

\begin{itemize}[leftmargin=*] 
    \item \textbf{Coarse-grained:}  Go to the green space area.
    
    \item \textbf{Medium-grained :} 
    Moving from the residential area to the green space area.

    \item \textbf{Fine-grained:} 
    A journey from a residential area to the green space area. Starting at timeslot 144 with a distance measuring 5.57 kilometers. 
    \textit{(Optional: It will travel through the following cells: [97, 103, 104, 105].)}

    \item  \textbf{Extremely-grained:}
    Route connecting green space and residential areas. The journey starts primarily in a green space/natural area (28.6\%). This is a mixed-use area with significant green space and residential functions, also featuring religious functions. 
    The area has good functional diversity, indicating high livability. and ends in a residential area (23.2\%). This area has a relatively singular function. The area has very high functional diversity, typical of urban centers.  
    \textit{(Note: Incorporates extremely rich semantic descriptions of the environment.)}
\end{itemize}
These multi-level descriptions enable \model to support a flexible spectrum of control, satisfying both open-ended simulation needs and precise path-planning requirements.
It is important to note that due to the model's exceptionally strong semantic understanding capabilities, more detailed instructions will reduce the diversity of the generated trajectories.
For example, using extremely detailed semantics causes the model to generate a unique representation for each area, resulting in routes that cover only the corresponding areas.
Therefore, we employ a fine-grained semantic representation, which is compatible with medium- and coarse-grained content.

\section{Details of \model}\label{app:detail_frame} 
This section provides detailed specifications for the components of our \model framework, complementing the descriptions in the main paper to ensure full reproducibility.

\subsection{Intention Encoder}
As discussed in the main paper, our instruction encoder is designed to transform natural language instructions into rich, token-level semantic embeddings. 
The encoder is built upon a pre-trained decoder-only Large Language Model (specifically, \texttt{Meta-Llama-3-8B})\footnote{https://github.com/McGill-NLP/llm2vec} and is fine-tuned to serve as a powerful text encoder using the three-stage process introduced in LLM2Vec \cite{llm2vec}.
The resulting encoder outputs a sequence of embeddings with dimension $d_{text}=4096$, which we linearly project to the model dimension $d$.

\subsection{MT-DiT Architectural Specifications}\label{app:prars_setting}
Our generative backbone, the Multi-modal Trajectory Diffusion Transformer (MT-DiT), is a custom Transformer architecture designed to effectively fuse trajectory and instruction information. It is composed of a stack of $N$ identical blocks.

\begin{itemize}[leftmargin=*]
    \item \textbf{Input Tokenization:} 
    The input trajectory $\boldsymbol{x}_t \in \mathbb{R}^{L \times 2}$ is first partitioned into non-overlapping patches of size $P$. These patches are flattened and linearly projected into the model's hidden dimension, $d$, to form the trajectory tokens. The instruction, encoded by LLM2Vec, is projected to the same dimension $d$. These two token sequences, along with the timestep embedding, form the input to the MT-DiT stack.

    \item  \textbf{Conditional Modulation (adaLN-Zero):} 
    We employ the \textbf{adaLN-Zero} mechanism for conditioning. The diffusion timestep embedding (and optional attribute embeddings) is processed by a specialized MLP to regress the scale ($\gamma$), shift ($\beta$), and dimension-wise scaling factor ($\alpha$) parameters. Crucially, the MLP is initialized to output zero for $\alpha$, effectively initializing each block as an identity function to stabilize training.

    \item  \textbf{Joint Attention for Fusion:} The modulated trajectory and instruction tokens are concatenated along the sequence dimension. This combined sequence is processed by a single multi-head self-attention layer. This is the critical step where deep, bidirectional fusion occurs, allowing trajectory tokens to attend to specific instruction tokens and vice versa.
    
    \item \textbf{Feed-Forward Networks:}
    After attention, the sequence is split back into its respective streams. Each stream is processed by a separate Feed-Forward Network (FFN) (GELU activation) to capture modality-specific features.

\end{itemize}

\begin{table}[h]
    \caption{General hyperparameters setting for \model.}
    \vspace{-3mm}
    \centering
    \begin{tabular}{lcrr} 
    \toprule
    Parameter & & Setting value & Refer range  \\ 
    \cmidrule(lr){1-4}
    Input Length & & 200 & 120 $\sim$ 256\\
    Patch size   & & 4 & 1 $\sim$ 8  \\
    
    Diffusion Steps & & 500  & 300 $\sim$ 500\\
    Sampling steps & & 100 & 50 $\sim$ 500\\
    $\beta$ (linear schedule) & & 0.0001 $\sim$ 0.02 & -- \\
    
    Embedding dim & & 384 & $\ge$ 128 \\
    Num heads & & 6 &  $\ge$ 2 \\
    
    MT-DiT blocks & & 6 & $\ge$ 4\\
    
    Batch size & & 1024 & $\ge$ 256 \\
    Parameter size & & 34.5 MB  & -- \\
     \bottomrule
    \end{tabular}
    \label{tab:model_para}
\end{table}

\noindent\textbf{MT-DiT Block.} 
The forward pass through a single MT-DiT block involves four key steps. First, the timestep embedding modulates the trajectory and instruction streams independently via adaLN. 
In addition, for some discrete attributes describing trajectory movements, we can treat them as optional embeddings, use the method in \cite{zhu2023difftraj} for embedding, and then add them to the time embedding.
Second, the two modulated token sequences are concatenated and fed into a shared Joint Attention layer, which is the critical site for deep, bidirectional fusion. 
Third, after the attention computation, the sequence is split back into its respective trajectory and instruction parts. 
Finally, each part is processed by a modality-specific FFN. Standard residual connections are used throughout the block.
After the final block, the instruction tokens are discarded. The processed trajectory tokens are passed through a final linear layer that projects them back to the flattened trajectory patch dimension to produce the final noise prediction.
A summary of the hyperparameters used in our model configuration is provided in Table \ref{tab:model_para}.

\subsection{TrajCLIP Configuration} \label{app:trajclip_detail}
To evaluate Instruction Faithfulness (RQ2), we trained a TrajCLIP model. The configuration details are as follows:
\begin{itemize}[leftmargin=*]
\item \textbf{Trajectory Encoder:} A standard Transformer Encoder with 6 layers, hidden dimension 256, and 8 attention heads.
\item \textbf{Text Encoder:} Frozen LLM2Vec encoder (same as the generator), and connected with a two-layer MLP to unifie the embedding dimension.
\item \textbf{Training:} The model was trained for 100 epochs using a contrastive loss with a learnable temperature parameter $\tau$. 
\end{itemize}

\subsection{Training and Inference Processes}

The end-to-end procedures for training and using \model are detailed in Algorithm \ref{alg:instraj_training} and Algorithm \ref{alg:instraj_inference}, respectively.

\noindent\textbf{Training.} As shown in Algorithm \ref{alg:instraj_training}, the training process follows the standard objective for diffusion models. In each step, we sample a ground-truth trajectory and its corresponding instruction from our training corpus. The instruction is encoded into a semantic embedding $\boldsymbol{c}$ using our frozen LLM2Vec encoder. We then corrupt the trajectory with Gaussian noise according to a randomly chosen timestep $t$. The MT-DiT model, $\boldsymbol{\epsilon}_\theta$, is then trained to predict the added noise by minimizing the mean-squared error, conditioned on both the noised trajectory $\boldsymbol{x}_t$, the timestep $t$, and the instruction embedding $\boldsymbol{c}$.

\begin{algorithm}[h]
\caption{Training Processes of \model}\label{alg:instraj_training}
\raggedright

\textbf{Training Process:}
\begin{algorithmic}[1]
\State \textbf{Require:} Training data $\mathcal{D} = \{(\boldsymbol{x}_{0}, I) \}$, trained Instruction Encoder $E_{\phi}$.
\Repeat
    \State Sample a trajectory-instruction pair $(\boldsymbol{x}_{0}, I) \sim \mathcal{D}$.
    \State Get instruction embedding $\boldsymbol{c} \leftarrow E_{\phi}(I) $.
    \State Sample a timestep  $t \sim \operatorname{Uniform}(\{1, \ldots, T\})$ and noise $\boldsymbol{\epsilon} \sim \mathcal{N}(\boldsymbol{0}, \mathbf{I})$.
    \State Create noised trajectory: $\boldsymbol{x}_t \leftarrow \sqrt{\bar{\alpha}_t}\boldsymbol{x}_{0} + \sqrt{1-\bar{\alpha}_t}\boldsymbol{\epsilon} $. 
    \State Take a gradient descent step on $\nabla_{\theta}\| \boldsymbol{\epsilon} - \boldsymbol{\epsilon}_\theta(\boldsymbol{x}_t, t, \boldsymbol{c})\|^2$. 
\Until{convergence}
\end{algorithmic}
\end{algorithm}
\vspace{-5mm}

\begin{algorithm}[h]
\caption{Inference Processes of \model}
\label{alg:instraj_inference}
\raggedright
\textbf{Inference Process:}
\begin{algorithmic}[1]
\State \textbf{Require:} Human travel instruction $I$, trained model $\boldsymbol{\epsilon}_\theta$, trained Instruction Encoder $E_{\phi}$.
\State Get instruction embedding $\boldsymbol{c} \leftarrow E_{\phi}(I)$.
\State Sample initial noise $\tilde{\boldsymbol{x}}_{T} \sim \mathcal{N}(\boldsymbol{0}, \mathbf{I})$.
\For{$t = T, \ldots, 1$}
    \State Sample $\boldsymbol{z} \sim \mathcal{N}(\boldsymbol{0}, \mathbf{I})$.
    \If{$t=1$} 
        \State $\boldsymbol{z} \leftarrow \boldsymbol{0}$.
    \EndIf
    \State Predict noise: $\hat{\boldsymbol{\epsilon}} \leftarrow \boldsymbol{\epsilon}_{\theta}(\tilde{\boldsymbol{x}}_{t}, t, \boldsymbol{c})$. 
    \State Denoise one step: $\tilde{\boldsymbol{x}}_{t-1} \leftarrow \frac{1}{\sqrt{\alpha_t}}(\tilde{\boldsymbol{x}}_{t}-\frac{1-\alpha_t}{\sqrt{1-\bar{\alpha}_t}}\hat{\boldsymbol{\epsilon}}) + \sigma_t\boldsymbol{z}$. 
\EndFor
\State \textbf{return} $\tilde{\boldsymbol{x}}_0$. 
\end{algorithmic}
\end{algorithm}

\noindent\textbf{Inference.} The inference process, detailed in Algorithm \ref{alg:instraj_inference}, reverses the noising procedure to generate a new trajectory. Given a new user instruction, we first encode it to obtain its conditioning embedding $\boldsymbol{c}$. Starting from a tensor of pure Gaussian noise $\tilde{\boldsymbol{x}}_T$, our trained model $\boldsymbol{\epsilon}_\theta$ iteratively denoises the sample for $T$ steps. At each step $t$, the model predicts the noise conditioned on the current sample $\tilde{\boldsymbol{x}}_t$ and the instruction $\boldsymbol{c}$, which is then used to compute the less-noisy sample $\tilde{\boldsymbol{x}}_{t-1}$. This process is repeated until a clean trajectory $\tilde{\boldsymbol{x}}_0$ is synthesized.

\section{Experiment and Setup}\label{app:exp_setup}

This section complements the main text by providing more details on our experimental setup, encompassing dataset statistics, detailed baseline configurations, and specific implementation choices.

\subsection{Dataset}\label{app:dataset}

We conduct our experiments on two large-scale, real-world public datasets: \textbf{Taxi-Chengdu} and \textbf{Taxi-Xi'an.}
Both datasets consist of taxi trajectories and are standard benchmarks in mobility research. Table \ref{tab:dataset} provides a statistical summary of these datasets after our pre-processing pipeline, which includes filtering out short or sparse trajectories and segmenting trips.
For consistency, we follow the standard protocol in \cite{zhu2023difftraj,zhu2024controltraj}: trajectories are normalized to a fixed length of $L=200$ points using linear interpolation.

\begin{table}[h]
    \caption{Statistics of two trajectory datasets.}
    \centering
    \small
    \begin{tabular}{lccc} 
    \toprule
    Dataset & Trajetory Number  & Area Number & Function Type   \\ 
    \cmidrule(lr){1-4}
    Chengdu  & \num{1095463}  & $50 \times 50$ &  19 \\
    Xi'an  & \num{1083784}  &  $50 \times 50$  & 17 \\
    \bottomrule
    \end{tabular}
    \label{tab:dataset}
\end{table}

\begin{figure}[h]
    \subfigure[Original trajectories]{
    \includegraphics[width=0.41\linewidth]{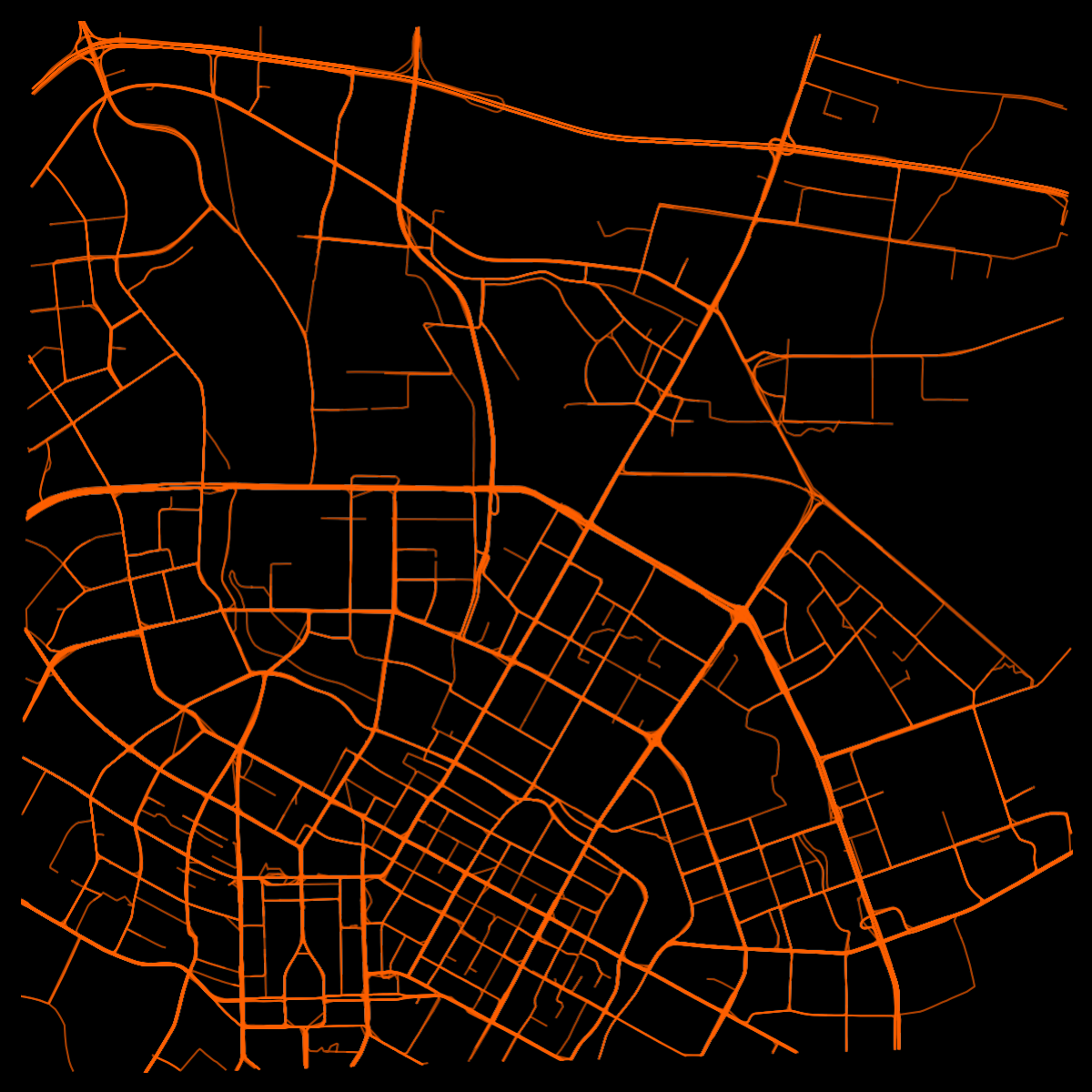}
    }\hspace{-0.02\linewidth} 
    \subfigure[Function map]{
    \includegraphics[width=0.525\linewidth]{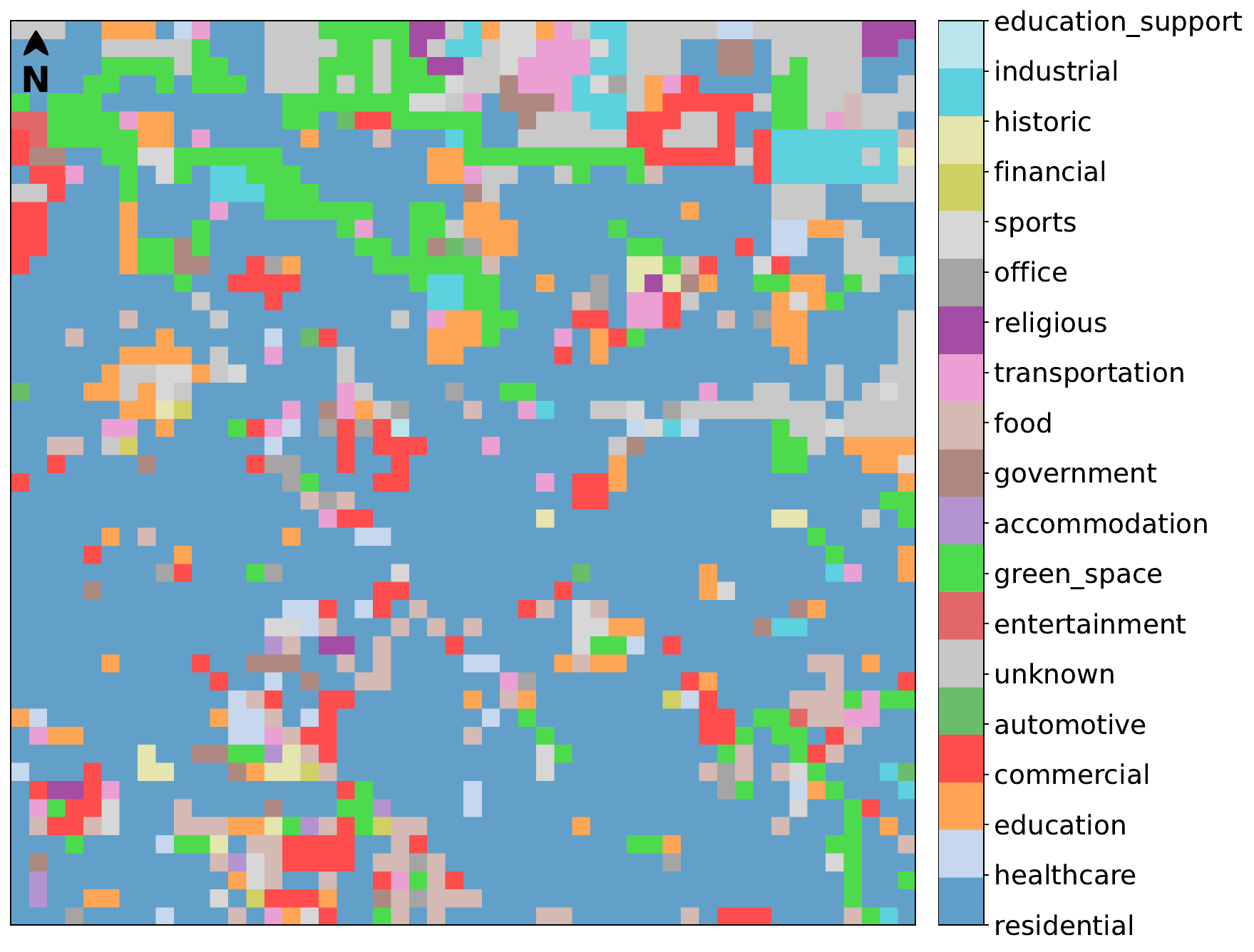}
    }
    \vspace{-3mm}
    \caption{Origin trajectory and function profile of Chengdu.}
    \Description[]{}
    \label{appfig:original_cd}
\end{figure}

\begin{figure}[h]
    \subfigure[Original trajectories]{
    \includegraphics[width=0.41\linewidth]{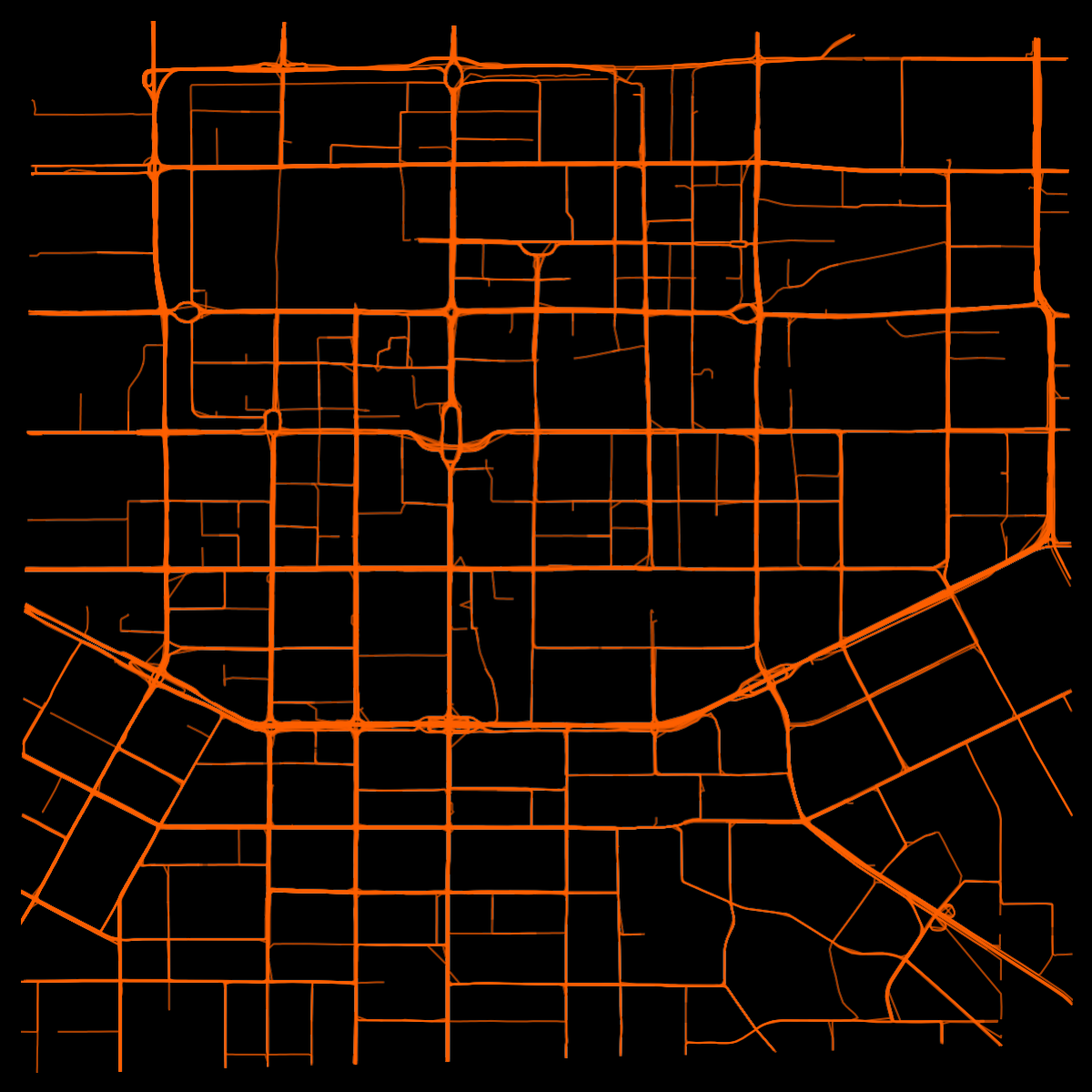}
    }\hspace{-0.02\linewidth} 
    \subfigure[Function map]{
    \includegraphics[width=0.525\linewidth]{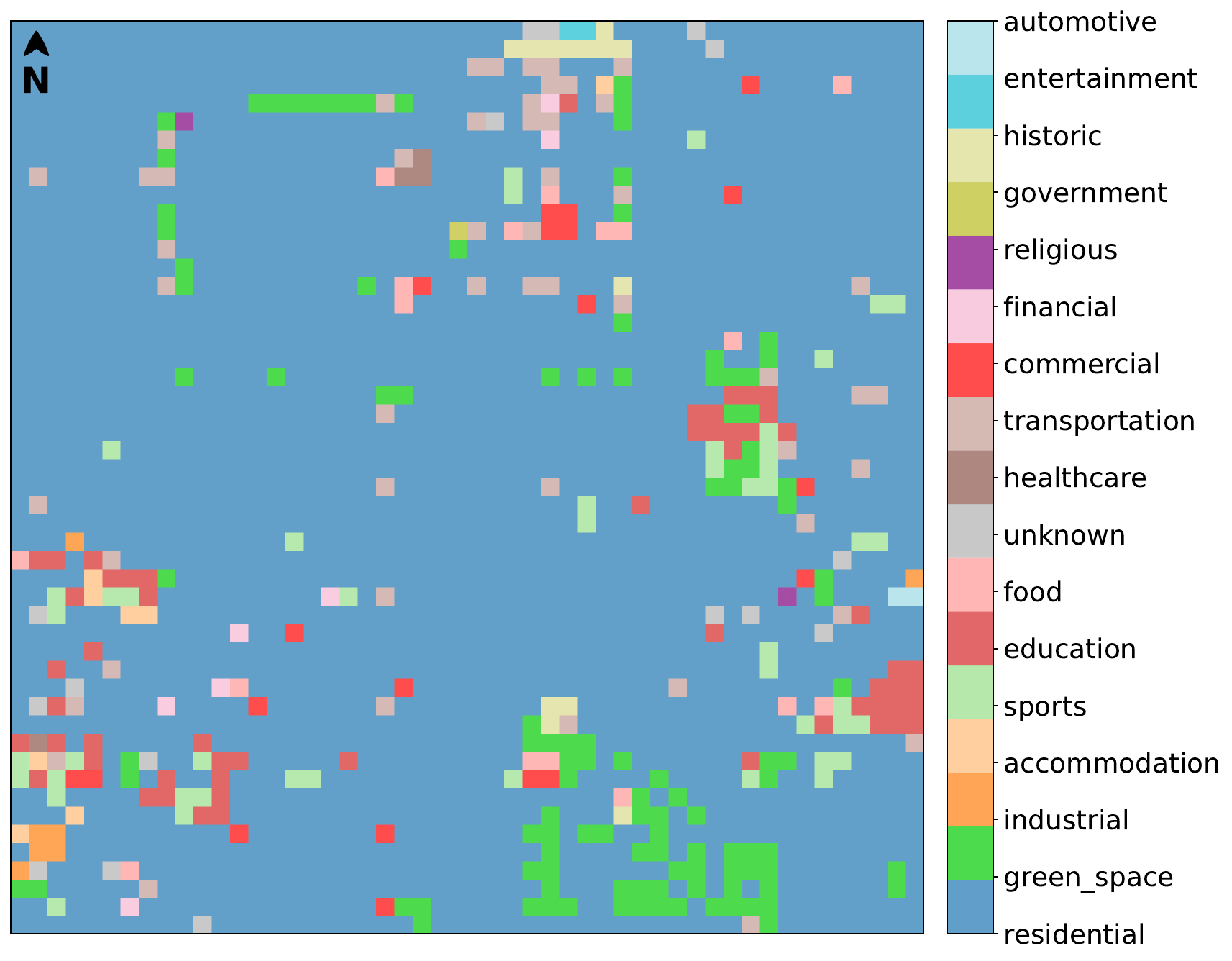}
    }
    \vspace{-3mm}
    \caption{Origin trajectory and function profile of Xi'an.}
    \Description[]{}
    \label{appfig:original_xa}
\end{figure}

To provide a more intuitive understanding of the urban environments, Figure \ref{appfig:original_cd} and Figure \ref{appfig:original_xa} present the spatial distribution of trajectories and the functional area layout for both cities. Subfigures (a) visualize the raw trajectory data, clearly outlining the primary road networks and traffic density patterns. Subfigures (b) display the corresponding functional maps constructed using Point-of-Interest (POI) data harvested from OpenStreetMap (OSM). In these visualizations, distinct colors represent different functional categories (e.g., Commercial, Residential, Recreational) obtained through our tag-to-function mapping process detailed in Section \ref{sec:distill_intention}. These visualizations underscore the distinct urban morphologies and provide the semantic foundation for our instruction generation process.

\subsection{Baseline Implementation Details}\label{app:baseline} 
This section outlines the specific architectural configurations of the baseline models. 
For Spatio-Temporal Fidelity (RQ1) and Efficiency (RQ3) experiments, we utilize the \textit{original, unmodified} implementations of all baselines to evaluate their native performance. For Instruction Faithfulness (RQ2), we explicitly modify DiffTraj and ControlTraj to accept semantic conditions, as detailed below.

\begin{itemize}[leftmargin=*] 

\item \textbf{VAE} \cite{chen2021trajvae}: Our implementation uses a symmetric encoder-decoder structure. The encoder utilizes a sequence of two 1D convolutional layers followed by a linear projection to define the latent space parameters. The decoder mirrors this architecture, employing a linear layer and two transposed 1D convolutional layers to reconstruct the trajectory. A kernel size of 4 is used throughout.

\item \textbf{TrajGAN \& DP-TrajGAN} \cite{zhang2022dp,rao2020lstm}: We employ an adversarial setup based on the TrajGAN paradigm. The generator is an LSTM-based network that transforms a noise vector (concatenated with conditions) into a trajectory sequence. The discriminator is an LSTM network tasked with distinguishing real trajectories from synthetic ones. For DP-TrajGAN, we evaluate it without the differential privacy noise injection to assess its generation capability as a standard GAN baseline.

\item \textbf{DiffWave} \cite{kongdiffwave}: We adapt the DiffWave architecture, originally designed for audio synthesis. Our implementation consists of a stack of 16 residual blocks. Each block features a bidirectional dilated 1D convolution, with outputs from the forward and backward paths gated by sigmoid and tanh activations.

\item \textbf{DiffTraj} \cite{zhu2023difftraj}: We follow the official implementation\footnote{https://github.com/Yasoz/DiffTraj}, utilizing a U-Net-like architecture with a ResNet backbone.
To adapt this model for instruction guidance, we modify its conditioning mechanism. The semantic embedding derived from our LLM2Vec encoder is projected and added to the timestep embedding, serving as the conditional input to the adaptive normalization layers.

\item \textbf{ControlTraj} \cite{zhu2024controltraj}: We use the official implementation\footnote{https://github.com/Yasoz/ControlTraj}, featuring a GeoUNet architecture designed to incorporate road network information. 
To integrate semantic guidance, we replace its original road segment embedding with our LLM2Vec-derived instruction embedding. This embedding is processed by the geo-attention layers, allowing for a direct comparison of geometric versus semantic guidance within the same architecture.

\item \textbf{\model(Bert) (Ablation)}: This variant isolates the contribution of the instruction encoder. The MT-DiT architecture remains identical to the full \model. However, the LLM2Vec encoder is replaced with a standard, pre-trained BERT model \cite{devlin2019bert}. The instruction embedding is obtained by mean-pooling the last hidden states of the BERT encoder.

\item \textbf{\model(DiT) (Ablation)}: This variant quantifies the benefits of our multi-modal fusion architecture. It uses our powerful LLM2Vec encoder but replaces the specialized MT-DiT backbone with a standard DiT architecture \cite{peebles2023scalable}. In this setup, token-level instruction embeddings are pooled into a single vector, added to the timestep embedding, and used to condition the DiT blocks via the standard adaLN-Zero mechanism. This mimics a global conditioning approach rather than our deep, token-level fusion.

\end{itemize}

\subsection{Evaluation Metrics}\label{app:metrics}
This section provides detailed formulations for the evaluation metrics used to assess model performance across spatio-temporal fidelity and instruction faithfulness.

\subsubsection{Spatio-Temporal Fidelity}
To assess the realism of the generated trajectories, we follow standard practices \cite{zhu2023difftraj} and quantify the distributional similarity between generated ($Q$) and real ($P$) trajectories using the Jensen-Shannon Divergence (JSD). We evaluate three key statistical attributes:
\begin{itemize}[leftmargin=*]
    \item \textbf{Spatial Density:} We partition the geographical area into a uniform grid and compute the probability distribution of trajectory points falling into each cell.

    \item \textbf{Trip Distribution:} We compute the joint probability distribution of the start and end locations to evaluate the fidelity of global patterns.

    \item \textbf{Relative Length:} We compute the distribution of the relative length with trajectory points.
\end{itemize}
For a given attribute, the JSD is calculated as:
\begin{equation}
JSD(P || Q) = \frac{1}{2} D_{KL}(P || M) + \frac{1}{2} D_{KL}(Q || M),
\end{equation}
where $M = \frac{1}{2}(P+Q)$  and $D_{KL}$ is the Kullback-Leibler divergence. A lower JSD score indicates higher fidelity.

\subsubsection{Instruction Faithfulness}
We use specific metrics to quantify adherence to semantic and spatial instructions.

\noindent\textbf{Functional Consistency Rate (FCR).} 
This metric evaluates whether a trajectory's start/destination points align with the functional areas specified in its instruction. 
For an instruction that requires a specific destination function (e.g., ``go to a commercial area''), we first identify the grid cell where the generated trajectory's endpoint is located. 
We then retrieve this cell's dominant functional type using our pre-computed city map  (from Section \ref{sec:distill_intention}), we identify the dominant functional category for the start cell $g_{\text{start}}$  and destination $g_{\text{dest}}$ of a generated trajectory. 
Let the instruction specify a required start function $f_{\text{start}}$ and destinated function $f_{\text{dest}}$. 
The FCR is the percentage of generated trajectories that satisfy both conditions:
\begin{align}
\text{FCR (S)} &= \frac{1}{|\mathcal{D}|} \sum_{i}^{|\mathcal{D}|} \mathbb{I}(f_{\text{start}}, g_{\text{start}}), \\
\text{FCR (D)} &= \frac{1}{|\mathcal{D}|} \sum_{i}^{|\mathcal{D}|} \mathbb{I}(f_{\text{dest}}, g_{\text{dest}}), 
\end{align}
where $\mathbb{I}(\cdot)$ is the indicator function. A higher FCR indicates better adherence to explicit spatial goals.

\noindent\textbf{Trajectory-Instruction Similarity (TIS).} 
To capture global semantic alignment beyond simple endpoints, we use the cosine similarity in the latent space of our TrajCLIP model (detailed in Section \ref{sec:LLM2vec}).
Given a generated trajectory $\boldsymbol{x}_{0}$ and instruction $I$, we first use the respective encoders of TrajCLIP to obtain their normalized embeddings, $\boldsymbol{v}_{\text{traj}}$ and $\boldsymbol{v}_{\text{text}}$. 
The TIS score is then calculated as the cosine similarity between these two vectors:
\begin{equation}
\text{TIS}(\boldsymbol{x}_0, I) = \frac{\boldsymbol{v}_{\text{traj}} \cdot \boldsymbol{v}_{\text{text}}^{T}}{||\boldsymbol{v}_{\text{traj}}|| \cdot ||\boldsymbol{v}_{\text{text}}||}.
\end{equation}
The final TIS reported in our experiments is the average score over the entire set of generated trajectories. 
A higher TIS score signifies a stronger semantic alignment between the generated path and the user's intent.

\begin{figure*}[t]
    \subfigure[DiffTraj]{
    \includegraphics[width=0.32\linewidth]{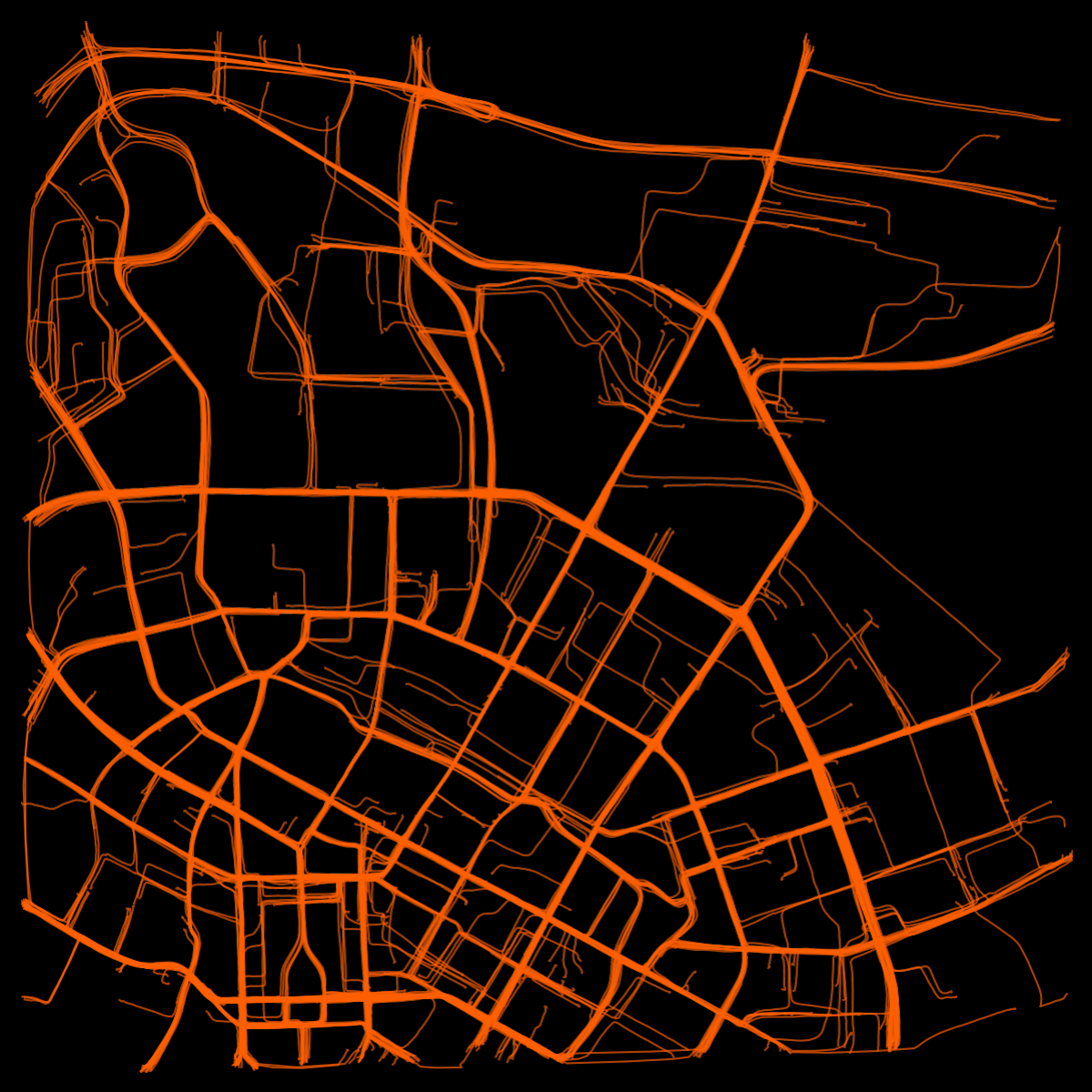} 
    }
    \subfigure[ControlTraj]{
    \includegraphics[width=0.32\linewidth]{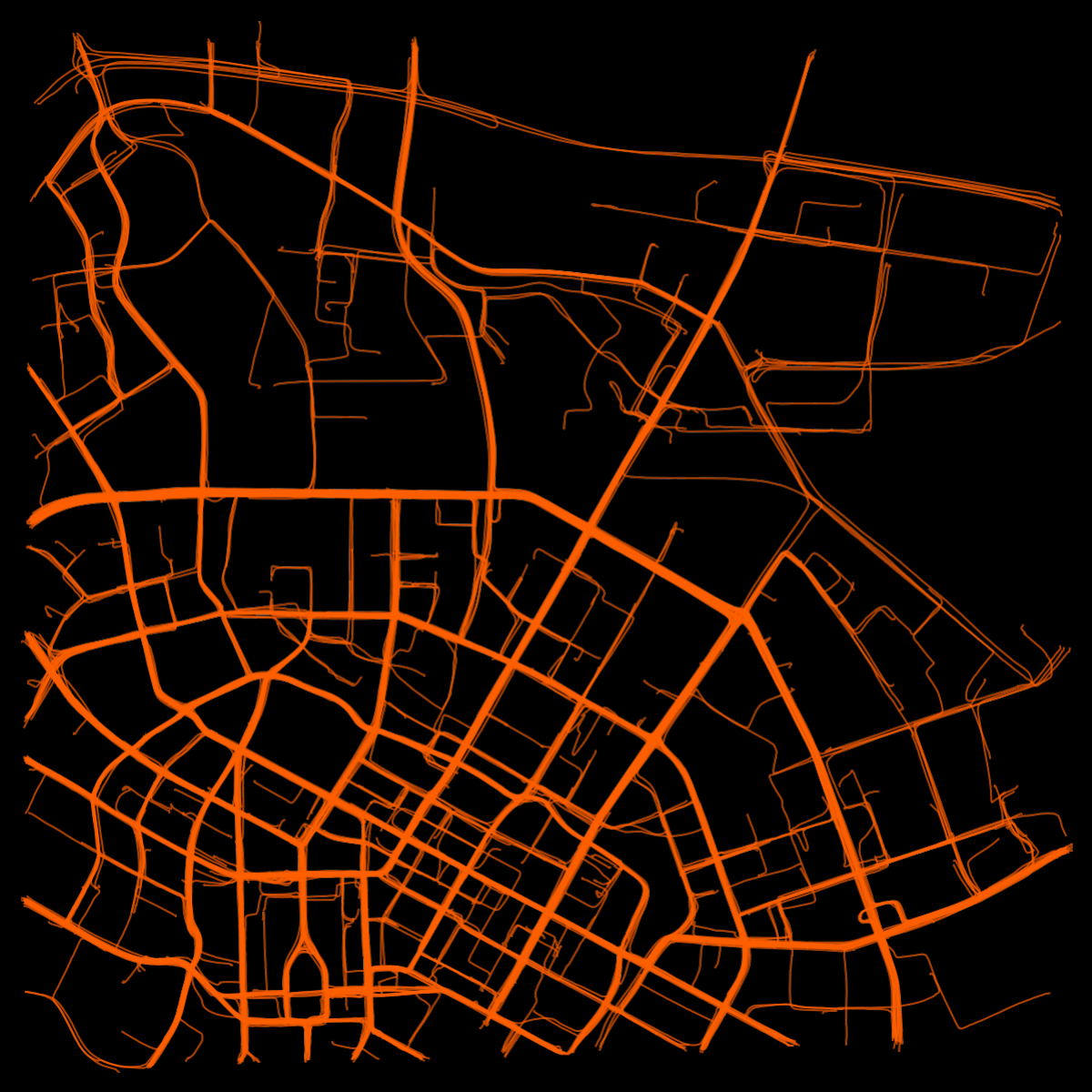}
    }
    \subfigure[\model]{
    \includegraphics[width=0.32\linewidth]{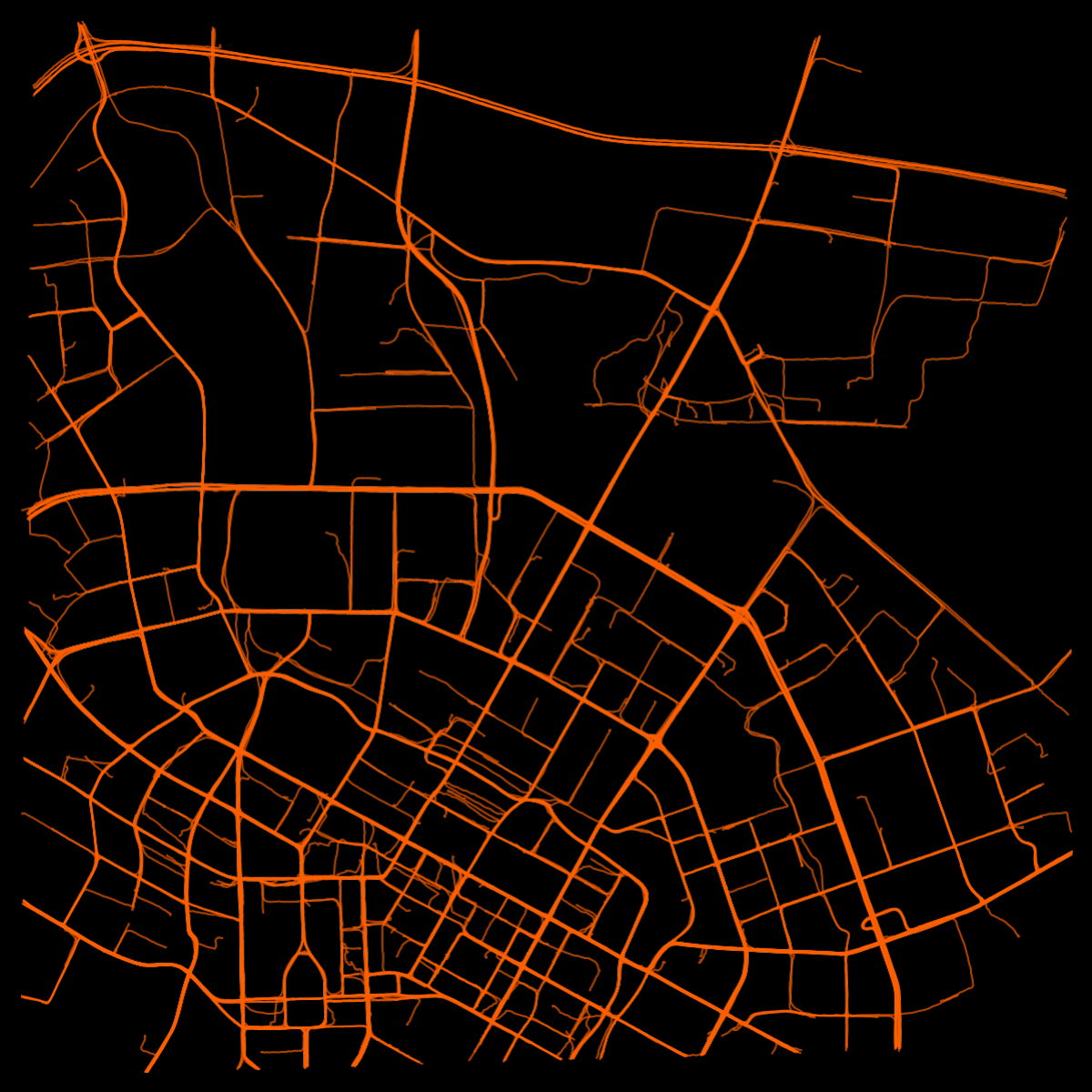}
    }
    \vspace{-5mm}
    \caption{Generated trajectory comparison of Chengdu.}
    \label{fig:large_vis_chengdu}
\end{figure*}

\begin{figure*}[t]
    \subfigure[DiffTraj]{
    \includegraphics[width=0.32\linewidth]{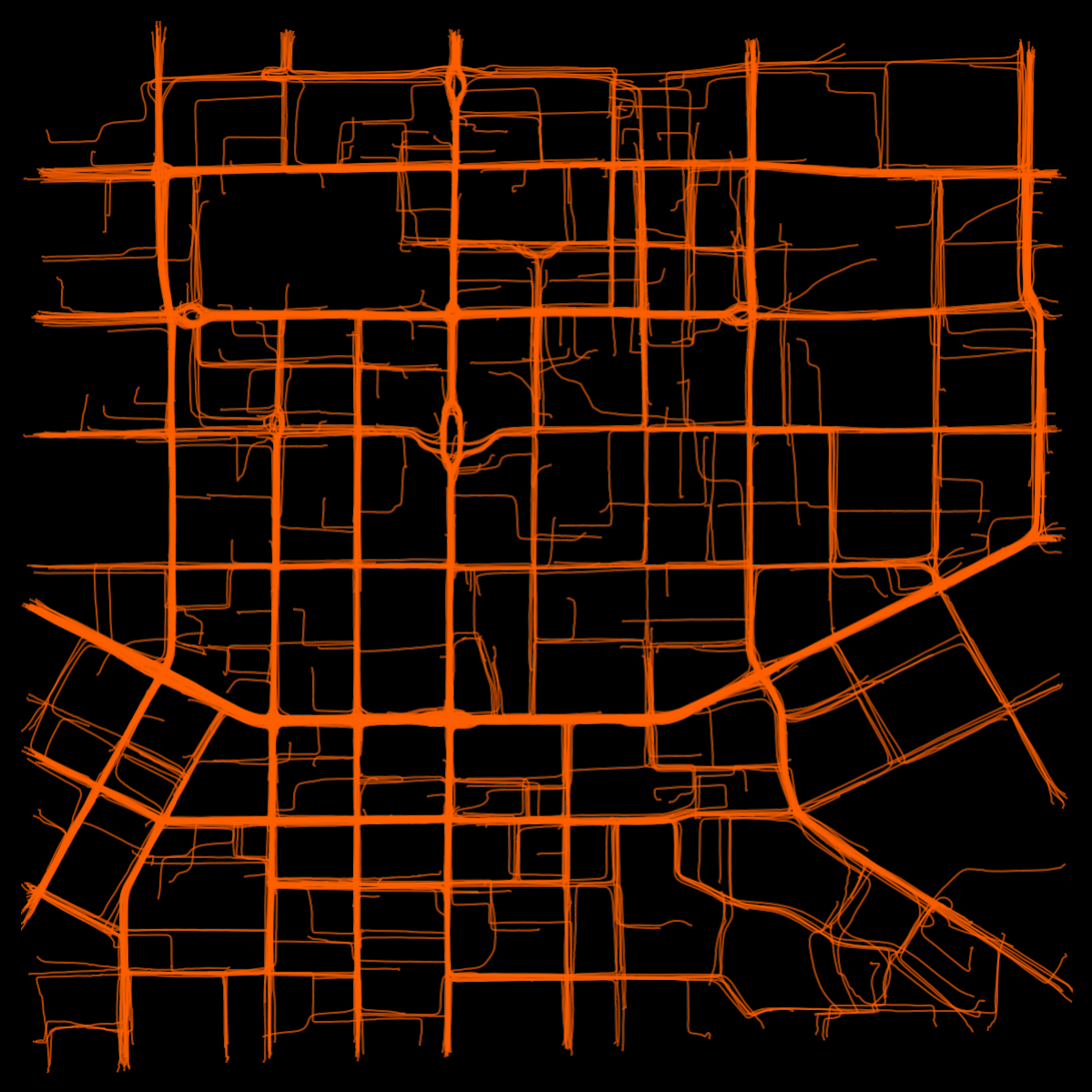} 
    }
    \subfigure[ControlTraj]{
    \includegraphics[width=0.32\linewidth]{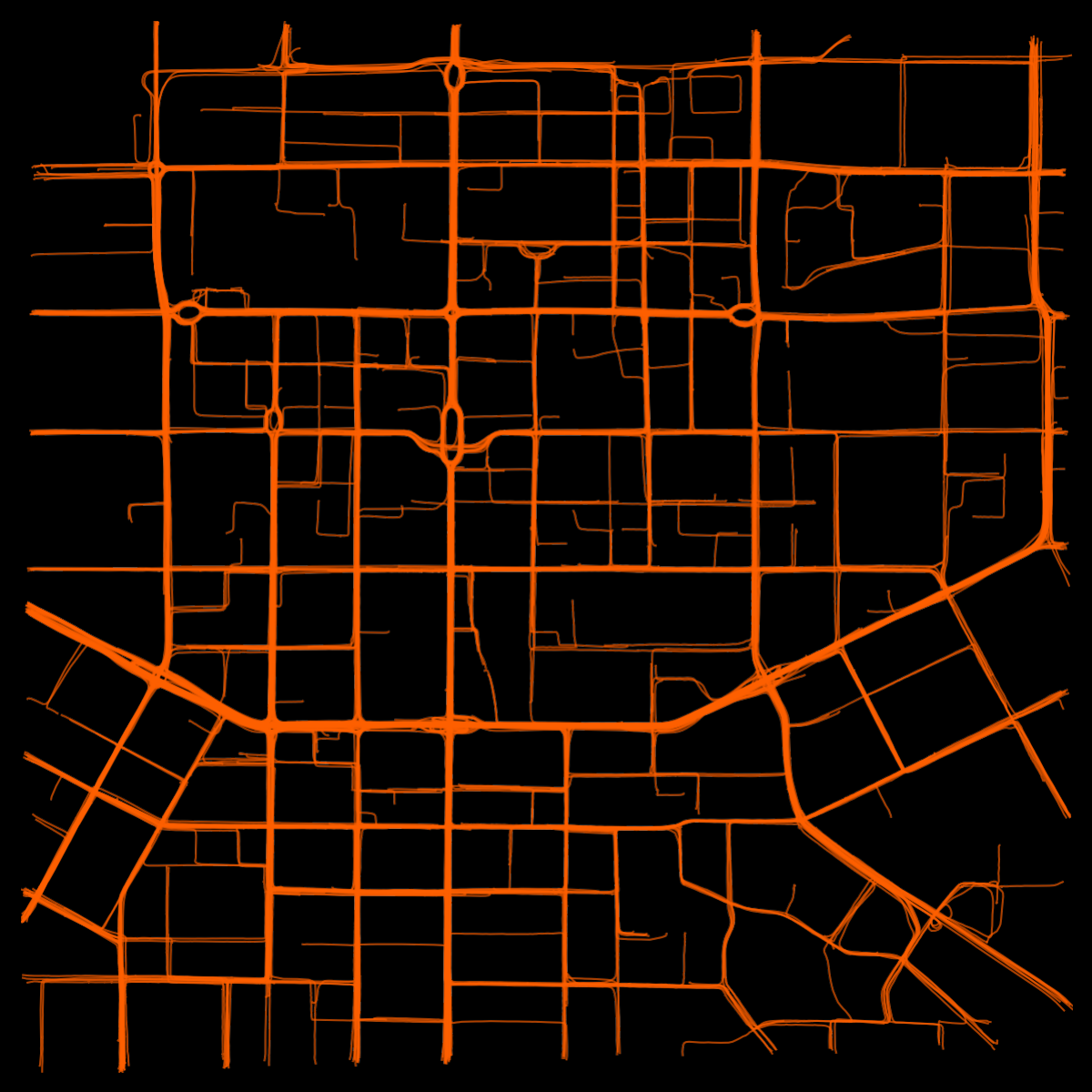}
    }
    \subfigure[\model]{
    \includegraphics[width=0.32\linewidth]{exp/instraj_xian.pdf}
    }
    \vspace{-5mm}
    \caption{Generated trajectory comparison of Xi'an.}
    \label{fig:large_vis_xian}
\end{figure*}

\section{Supplementary Experiments}
To provide a more holistic evaluation of \model, we conducted additional experiments focusing on cross-city generalizability, robustness to linguistic variations, and the impact of instruction granularity.

\subsection{Cross‑city Generalizability}\label{app:generalizability}
A critical challenge in mobility generation is the ability to generalize across different urban environments with distinct road network topologies and coordinate systems. To evaluate the transferability of \model, we conducted a transfer learning experiment where the model was pre-trained on the Chengdu dataset and fine-tuned on the Xi'an dataset using varying ratios of the target training data (5\%, 10\%, and 50\%).
\textbf{Train from Scratch:} The model is initialized randomly and trained solely on the limited subset of the Xi'an dataset.
\textbf{Train with Transfer:} The model inherits weights from the Chengdu-pretrained model and is fine-tuned on the Xi'an subset.

\begin{table}[h]
\centering
\small
\caption{Cross-city transfer performance (scratch/transfer).}
\vspace{-3mm}
\begin{tabular}{l|ccc}
\toprule
\textbf{Data Ratio} & \textbf{Density $\downarrow$} & \textbf{Trip $\downarrow$} & \textbf{Length $\downarrow$} \\
\midrule
5\% & 0.0268 / 0.0214	& 0.0395 / 0.0258 & 0.0358 / 0.0198 \\
10\% & 0.0201 / 0.0185 & 0.0283 / 0.0244 & 0.0337 / 0.0182 \\
50\% & 0.0115 / 0.0097 & 0.0145 / 0.0131 & 0.0209 / 0.0165  \\
100\% & 0.0093 & 0.0110 & 0.0131 \\
\bottomrule
\end{tabular}
\label{tab:cross_city_transfer}
\end{table}

As presented in Table \ref{tab:cross_city_transfer}, the results demonstrate the strong transferability of \model. 
In the few-shot setting (5\% data), the pre-trained model significantly outperforms the scratch baseline (e.g., reducing Trip JSD error), suggesting that the learned semantic mobility patterns are universal and transfer well across diverse urban topologies. 
Furthermore, with only 50\% of the target data, the transfer model rapidly converges to a performance level comparable to the fully-trained reference (Density JSD: 0.0097 vs. 0.0093). This confirms that \model offers high data efficiency, making it a viable solution for cities where large-scale trajectory data is scarce.

\subsection{Visualization}\label{app:supp_vis}
Figure \ref{fig:large_vis_chengdu} and Figure \ref{fig:large_vis_xian} display the macro-level spatial distribution of a large sample of trajectories generated by \model and our strongest baseline, DiffTraj and ControlTraj, across the entire urban areas of Chengdu and Xi'an, respectively.
As can be observed, while both models successfully outline the main arterial roads, the trajectories generated by \model exhibit a higher degree of realism and adherence to the underlying road network, particularly in complex intersections and less-transited areas. 
Furthermore, the overall traffic flow patterns generated by \model appear more natural and closely resemble the ground-truth distributions shown in Figure \ref{appfig:original_cd} and Figure \ref{appfig:original_xa}. These qualitative results provide strong visual evidence that corroborates the quantitative superiority of \model in spatio-temporal fidelity.

\subsection{Instruction Quality \& Granularity Analysis}\label{app:ins_ana}

\noindent\textbf{Impact of Instruction Quality.} 
To justify the necessity of our LLM-based instruction refinement pipeline, we evaluated the model's robustness against linguistic perturbations. We compared three instruction generation strategies: 
Template-only (e.g., Go from [Residential] to [Commercial]), Template with Synonym replaced (e.g., changing Go to'' to Head towards'' or Commute to''), LLM-refined (using LLM to rephrase the instructions)

Table \ref{tab:instruction_quality} evaluates the model's robustness against linguistic perturbations. 
Although the template-only approach achieved a high FCR score, it lacks linguistic diversity. When we introduced lexical variation (synonym replacement or LLM modification), its FCR metric declined further, yet it still lacked linguistic diversity. This indicates overfitting to specific keywords.
In contrast, our \textit{LLM-refined} approach demonstrates superior robustness, achieving the highest TIS score (0.371) and maintaining high consistency despite linguistic variations, effectively capturing the underlying semantic intent.
This proves that training on LLM-refined data prevents the model from overfitting to rigid sentence structures, significantly enhancing its robustness in real-world interactions.

\begin{table}[h]
\centering
\small
\caption{Impact of instruction quality on faithfulness.}
\vspace{-3mm}
\begin{tabular}{l|ccc}
\toprule
\textbf{Instruction Type} & \textbf{FCR-S $\uparrow$} & \textbf{FCR-D $\uparrow$} & \textbf{TIS $\uparrow$} \\
\midrule
Template-only & 83.35 & 82.68 & 0.369 \\
Template (Synonym replaced) & 80.18 & 79.42 & 0.327 \\
LLM-refined (Ours) & 83.03 & 82.54 & 0.371 \\
\bottomrule
\end{tabular}
\label{tab:instruction_quality}
\end{table}

\noindent\textbf{Analysis of Instruction Granularity.} 
We further explored how the granularity of instructions affects generation. 
As shown in \ref{tab:instruction_granularity}, we observed a distinct trade-off: Fine-grained instructions (e.g., specifying time, OD, and purpose) yield the highest TIS and FCR scores, as they constrain the solution space. 
Coarse-grained instructions (e.g., vague purpose only), while resulting in slightly lower consistency scores, produce a higher diversity of trajectories. This indicates that \model correctly interprets the strictness of a command: behaving deterministically when constraints are tight, and stochastically when the user's intent is open-ended.

\begin{table}[h]
\centering
\small
\caption{Impact of instruction granularity on faithfulness.}
\vspace{-3mm}
\begin{tabular}{l|ccc}
\toprule
\textbf{Instruction Type} & \textbf{FCR-S $\uparrow$} & \textbf{FCR-D $\uparrow$} & \textbf{TIS $\uparrow$} \\
\midrule
Coarse-grained & -- & 68.76 & 0.254 \\
Medium-grained & 75.32 & 73.92 & 0.336 \\
Fine-grained & 83.03 & 82.54 & 0.371 \\
\bottomrule
\end{tabular}
\label{tab:instruction_granularity}
\end{table}

\subsection{OD‑based Analysis}\label{app:odanalysis}
To strictly validate the model's capability to handle precise spatial constraints (as qualitatively demonstrated in the ``Designated Area'' case study), we evaluated the quantitative accuracy of the generated trajectories in hitting the instructed Origin and Destination (OD) regions. 
We define \textit{Accuracy} as the percentage of trajectories where the generated start/end points fall strictly within the bounding boxes specified in the instruction. 
As reported in Table \ref{tab:od_acc}, \model demonstrates exceptional precision across both datasets, achieving Origin Accuracy scores exceeding 98\% and Destination Accuracy scores surpassing 96\%. These near-perfect hit rates confirm that \model does not merely generate plausible patterns but can also serve as a reliable tool for controllable simulation tasks requiring strict adherence to geometric constraints.

\begin{table}[h] 
\centering \small \caption{Quantitative accuracy of OD constraint satisfaction.} 
\vspace{-3mm} 
\begin{tabular}{l|cc} 
\toprule 
\textbf{Dataset} & \textbf{Origin Accuracy} & \textbf{Destination Accuracy} \\ 
\midrule 
    Chengdu & 98.3 & 96.7 \\ 
    Xi'an & 99.2 & 97.1 \\ 
\bottomrule 
\end{tabular} 
\label{tab:od_acc} 
\vspace{-3mm} 
\end{table}

\end{appendix}
\end{document}